\documentclass[10pt]{article}
\usepackage[top=0.85in,left=0.75in,footskip=0.75in,marginparwidth=2in]{geometry}

\usepackage[utf8]{inputenc}
\usepackage{booktabs}
\usepackage{cite}
\usepackage{longtable}
\usepackage{afterpage}
\usepackage{subcaption}
\usepackage{booktabs}
\usepackage{hhline}
\usepackage{rotating}
\usepackage{placeins}
\usepackage{multirow}
\usepackage{comment}
\usepackage{graphicx,wrapfig,lipsum}
\usepackage{float}

\usepackage{nameref,hyperref}

\usepackage[pagewise]{lineno}

\usepackage{microtype}
\DisableLigatures[f]{encoding = *, family = * }

\usepackage{changepage}

\usepackage[aboveskip=1pt,labelfont=bf,labelsep=period,singlelinecheck=off]{caption}

\makeatletter
\renewcommand{\@biblabel}[1]{\quad#1.}
\makeatother

\usepackage{lastpage,fancyhdr,graphicx}
\usepackage{epstopdf}
\fancyheadoffset[L]{2.25in}
\fancyfootoffset[L]{2.25in}

\usepackage{color}

\definecolor{Gray}{gray}{.25}

\usepackage{graphicx}
\usepackage{adjustbox}

\usepackage{sidecap}

\usepackage{wrapfig}
\usepackage[pscoord]{eso-pic}
\usepackage[fulladjust]{marginnote}
\reversemarginpar

\begin{document}
\vspace*{0.35in}

\begin{flushleft}
{\Huge
\textbf\newline{The role of System~1 and System~2 semantic memory structure in human and LLM biases}}

\bigskip

\textbf{Katherine Abramski$^{1}$, Giulio Rossetti$^{2,+}$, Massimo Stella$^{3,+}$}
\newline
\\
1 - Department of Computer Science, University of Pisa, Italy
\\
2 - Institute of Information Science and Technologies - National Research Council, Italy
\\
3 - CogNosco Lab, Department of Psychology and Cognitive Science, University of Trento, Italy 
\\
+ - These authors contributed equally
\newline
\\
\textbf{Corresponding author: katherine.abramski@phd.unipi.it}
\bigskip

\end{flushleft}

\bigskip

\textbf{Keywords:} dual process theory, implicit bias, stereotype, semantic network, generative AI, social cognition, conceptual knowledge, multilayer network

\bigskip



{\small
\noindent Implicit biases in both humans and large language models (LLMs) pose significant societal risks. Dual process theories propose that biases arise primarily from associative System~1 thinking, while deliberative System~2 thinking mitigates bias, but the cognitive mechanisms that give rise to this phenomenon remain poorly understood. To better understand what underlies this duality in humans, and possibly in LLMs, we model System~1 and System~2 thinking as semantic memory networks with distinct structures, built from comparable datasets generated by both humans and LLMs. We then investigate how these distinct semantic memory structures relate to implicit gender bias using network-based evaluation metrics. We find that semantic memory structures are irreducible only in humans, suggesting that LLMs lack certain types of human-like conceptual knowledge. Moreover, semantic memory structure relates consistently to implicit bias only in humans, with lower levels of bias in System~2 structures. These findings suggest that certain types of conceptual knowledge contribute to bias regulation in humans, but not in LLMs, highlighting fundamental differences between human and machine cognition.
}

\section{Introduction}
Implicit biases can have significant societal consequences, often manifesting in the form of stereotypes that negatively impact certain groups. A large body of work has shown that such biases often arise automatically and outside conscious awareness \cite{greenwald1995implicit, greenwald1998measuring}, making them particularly challenging to regulate \cite{devine2012long}. Similar concerns have emerged around large language models (LLMs), which have been shown to exhibit harmful implicit biases despite performing well on explicit bias benchmarks \cite{bai2025explicitly,zhao2024comparative}. This parallels the human tendency for implicit attitudes to diverge from explicit beliefs \cite{perugini2005predictive, nosek2007pervasiveness,  monteith2001taking}, raising questions about whether humans and LLMs share similar cognitive architectures, and sparking renewed interest in the cognitive mechanisms underlying biased reasoning in humans and machines alike \cite{hagendorff2023human,yax2024studying, kamruzzaman2024prompting}.

A dominant framework for understanding the nature of implicit bias in humans is provided by dual process theories of cognition \cite{strack2004reflective, gawronski2006associative, gawronski2011associative}, which distinguish between two interacting systems of thought: a fast, automatic System~1, and a slower, deliberative System~2 \cite{sloman1996empirical, stanovich1999rational, kahneman2011thinking}. System~1 relies heavily on learned associations, enabling efficient pattern recognition but making it prone to systematic errors when associations reflect misleading correlations or social stereotypes. Instead, System~2 employs logical reasoning, abstract thinking, and cognitive control, and can monitor, evaluate, and sometimes override System~1 responses \cite{payne2005conceptualizing, kahneman2011thinking}. Thus, implicit biases are typically attributed to System~1, while bias regulation depends on the selective engagement of System~2 \cite{devine1989stereotypes, monteith1993self, devine2012long}. While there is extensive behavioral evidence supporting this functional distinction, \cite{stanovich1999rational, evans2010intuition}, much of this work focuses on characterizing differences in performance, speed, or controllability \cite{kahneman2011thinking, bargh1999unbearable}, rather than providing a mechanistic account of \textit{how} certain representational structures and processes contribute to systematic functional differences \cite{evans2013dual, moors2006problems, bursell2021we}. Clarifying this gap requires moving beyond functional characterizations and asking how different forms of knowledge organization bring about observable differences in biased-related behavior.

Several theoretical accounts have attempted to address this gap by arguing for the existence of two memory systems: an associative system and a rule-based system \cite{sloman1996empirical, smith2000dual, evans2008dual}, aligning with System~1 and System~2, respectively. Associative systems are characterized by statistical regularities, pattern completion, and similarity-based organization \cite{anderson1983spreading, collins1975spreading}, whereas rule-based systems are defined by logical rules, symbolic operations, and propositional (i.e. declarative) knowledge \cite{simon1971human, fodor1975language}. While these accounts aim to provide a mechanistic explanation of the dual process divide, they face two persistent limitations. First, they often blur the distinction between representational structure and process \cite{castro2020contributions, hills2022mind}. Associative systems are typically specified both in terms of their structural organization and their processing dynamics \cite{collins1969retrieval,collins1975spreading}, whereas descriptions of rule-based systems focus primarily on processing dynamics in terms of reasoning procedures, while leaving the structure of the underlying knowledge representations underspecified \cite{sloman1996empirical,reder1980partial, anderson1998integrated}. As a result, it remains unclear whether systematic functional differences are driven by differences in memory structure, processing mechanisms, or interactions between the two \cite{strack2004reflective, gawronski2011associative, de2021attitudes}. Second, these accounts are often abstract, providing limited guidance for how representational differences might be operationalized, measured, or compared empirically \cite{evans2008dual,evans2011dual,evans2013dual}.

\subsection{Modeling conceptual knowledge with semantic memory networks}
One way to address this gap is to focus explicitly on representational structure by modeling the different forms of conceptual knowledge associated with the two memory systems. Semantic memory networks provide a natural framework for this goal \cite{collins1969retrieval, collins1975spreading, barr1981handbook, woods1975s}. Associative memory has long been represented as networks of concepts connected by associative links often built from free associations \cite{collins1975spreading, anderson1983spreading, castro2020contributions, siew2019cognitive}. Such representations have been particularly successful in capturing properties commonly attributed to System~1 thinking, including similarity-based organization and lexical retrieval \cite{collins1975spreading, de2019small}. In contrast, representing rule-based memory in network form poses a greater challenge, since it requires flattening logical rules, symbolic operations and propositional knowledge onto an explicit memory structure \cite{sloman1996empirical, reder1980partial, anderson1998integrated}. Nevertheless, several lines of research suggest that relational network representations -- such as feature-based or categorical networks -- can preserve important aspects of rule-based and propositional knowledge \cite{collins1969retrieval, smith1974structure, mcrae2005semantic, anderson1997act}. Feature-based networks represent concepts in terms of their defining or characteristic features \cite{smith1974structure, mcrae2005semantic}, and can be built from feature norms or dictionary definitions by connecting concepts to their defining characteristics \cite{vincent2016latent, kozima1996similarity}. Categorical networks encode taxonomic relations and logical category constraints that reflect category inclusion \cite{collins1969retrieval, steyvers2005large}, and are commonly constructed using large-scale lexical resources such as WordNet \cite{miller1995wordnet}, which link concepts through typed semantic relations such as synonymy, antonymy, hypernymy, and hyponymy. Crucially, although representing rule-based memory in relational network form abstracts away from symbolic operations and logical rules, it preserves the core relational knowledge that underlies deliberative logical reasoning \cite{brachman1979epistemological, sowa1983generating}, providing a structural foundation for System~2 thinking \cite{smith1974structure, kintsch1998comprehension}. Thus, by encoding different forms of conceptual knowledge in semantic memory networks, associative and rule-based memory systems are put on the same representational plane. Moreover, the network representations enable comparative empirical investigations which can be carried out using well-studied tools, frameworks, and methodologies designed specifically for network models \cite{steyvers2005large}.

One such representational framework that is particular well-suited for this problem is multilayer networks, which allow multiple types of relationships between the same set of nodes to be represented as distinct but interconnected network layers \cite{boccaletti2014structure, kivela2014multilayer}. Unlike traditional single-layer networks, which collapse different relationship types into a single structure, multilayer networks preserve the distinct organization of each type of relationship while still allowing them to be analyzed within a unified system \cite{kivela2014multilayer, de2013mathematical}. Multilayer approaches have therefore become increasingly common in the study of complex cognitive systems that involve several types of complex relationships, including the study of language learning \cite{citraro2023feature} and the mental lexicon \cite{stella2024cognitive}. Within this framework, different forms of conceptual knowledge -- associative, feature-based, and categorical -- can be modeled as separate layers of a multilayer semantic network. Representing knowledge in this way allows the structure of each knowledge type to be analyzed independently while still considering their relationship within a broader conceptual system. Importantly, this approach allows us to focus explicitly on representational structure, thereby avoiding the conflation of structure and cognitive process that characterizes many dual-process accounts.

Another advantage of representing conceptual knowledge in the form of semantic memory networks is that it enables the use of well-established cognitive tools for investigating behavioral phenomena within network structures. Specifically, spreading activation is a widely used framework that simulates a lexical search process within memory networks \cite{collins1975spreading, anderson1983spreading}, and it has been applied extensively in psychology and linguistics. Recently, this framework has been applied for developing a tool for evaluating biases in word association networks \cite{abramski2025word}. It works by using spreading activation within a network to simulate semantic priming, which then produces a metric that measures the strength of association between words, which is used as a measure of bias.

Together, multilayer network modeling and spreading activation dynamics provide a principled framework for linking the structure of conceptual knowledge to observable patterns of biased association. In the present study, we model associative and rule-based memory systems as multilayer networks that represent different types of conceptual knowledge, and we apply the aforementioned bias evaluation methodology using spreading activation to investigate the relationship between bias and semantic memory structure. This approach addresses a significant research gap in two ways. First, it isolates representational structure as the primary object of study, helping to disentangle structural differences in conceptual knowledge from differences in cognitive processing. Second, it provides a concrete empirical framework for investigating how different knowledge structures give rise to bias-related behavior.

\subsection{Humans and LLMs: a shared framework}
Applying this framework to LLMs in addition to humans offers an opportunity to deepen our understanding of the cognitive mechanisms underlying biased reasoning. As LLMs continue to advance, their behavior increasingly resembles that of humans, making the dual process framework a useful metaphor for describing, studying, and potentially improving their capabilities \cite{hagendorff2023human, li2025system, ziabari2025reasoning, yang2025llm2}. Owing to their connectionist architecture, LLMs can be considered System~1 machines by design \cite{rumelhart1986parallel, smolensky1988proper}. Like human associative processes, these models rely primarily on distributed statistical representations learned from large corpora. As a result, they are prone to many of the same systematic reasoning errors observed in human cognition, including harmful biases and stereotypes \cite{acerbi2023large, bender2021dangers, garimella2021he, de2026cognitive}. Unlike humans, however, LLMs lack an explicit System~2-like symbolic architecture that could help regulate such biases through structured reasoning or propositional knowledge. This limitation has motivated efforts to enhance language models with more structured forms of conceptual knowledge intended to support reasoning and bias regulation. In particular, several approaches have attempted to introduce System~2-like information by incorporating dictionary definitions \cite{an2022learning, kaneko2021dictionary}, knowledge graphs \cite{mruthyunjaya2023rethinking, zheng2023does, kumar2025detecting}, or other forms of structured semantic knowledge \cite{ma2024debiasing, agrawal2024can}. These approaches are motivated by the idea that structured conceptual representations—similar to those associated with rule-based memory in humans—may help constrain or refine purely associative representations. At the same time, more recent models have demonstrated improved performance on logical reasoning tasks \cite{xiang2025towards, de2024system, hagendorff2023human}, as well as stronger results on explicit bias benchmarks \cite{bai2025explicitly, zhao2024comparative}, even without incorporating explicit System~2-like reasoning mechanisms. This raises questions about whether bias regulation in LLMs relies on mechanisms fundamentally different from those in humans \cite{brady2025dual}, and more broadly, whether deliberative reasoning based on propositional knowledge can emerge from a purely associative architecture \cite{pavlick2023symbols,wei2022chain, wei2022emergent}. These questions echo long-standing debates in cognitive science and AI about the relationship between associative and rule-based memory systems \cite{sloman1996empirical, fodor1988connectionism, smith2000dual}, and highlight the value of comparative approaches to studying how dual process cognition relates to bias in humans and LLMs. these developments highlight the value of comparative approaches that examine humans and artificial systems within a shared conceptual framework. Studying how different forms of conceptual knowledge relate to biased reasoning across humans and LLMs may therefore help clarify both the cognitive mechanisms underlying human bias and the representational limitations of current language models \cite{brady2025dual, hagendorff2023human}, informing the development of less biased artificial systems. To investigate these questions, we examine how different forms of conceptual knowledge relate to implicit bias in both humans and language models.

\subsection{The present study}
In the present study, we adopt this comparative approach to investigate how different semantic memory structures relate to implicit bias in humans and LLMs. Using comparable datasets generated by humans and LLMs, we model System~1 and System~2 thinking as multilayer semantic memory networks that encode different types of conceptual knowledge. Specifically, System~1 is modeled as an associative network layer constructed from free associations. System~2, in contrast, is represented using two complementary network layers that capture structured forms of conceptual knowledge: a feature-based network built from dictionary definitions and a categorical network built from categorical (taxonomic) relations. This representation reflects the fact that rule-based memory encompasses multiple forms of structured knowledge rather than a single representational format. These multilayer networks are constructed in parallel for humans and for two widely used open-source LLMs: Mistral (mistral-7b) and Llama3 (llama3.1-8b).

We then conduct two sets of analyses. First, we examine the structural reducibility of the multilayer networks in order to determine whether the different forms of conceptual knowledge represented in each layer constitute empirically distinct structures. We find that they are irreducible in humans but not in LLMs, suggesting that LLMs lack certain human-like conceptual knowledge, specifically, taxonomic knowledge. Second, we investigate how semantic memory structure relates to gender stereotypes by applying the spreading-activation–based bias evaluation framework described earlier. We find that semantic memory structure relates systematically to implicit bias in humans, with reduced bias in System~2-like structures, but inconsistently in both Mistral and Llama3. Together these findings highlight fundamental differences between human and machine cognition, suggesting that the mechanisms that contribute to regulating bias in humans and LLMs are fundamentally different.

\section{Materials and Methods}

This section describes the construction of the multilayer semantic networks and the analytical methods used in the study. We first outline how the networks were built from datasets generated by humans and two LLMs: Mistral (mistral-7b) and Llama3 (llama3.1-8b). We then describe the structural reducibility analysis used to evaluate whether the network layers encode distinct forms of conceptual knowledge. Finally, we present the methodology used to quantify gender stereotypes within each network layer.

\subsection*{Multilayer network construction}
\begin{figure}[h]
\centering
\includegraphics[width=.5\linewidth]{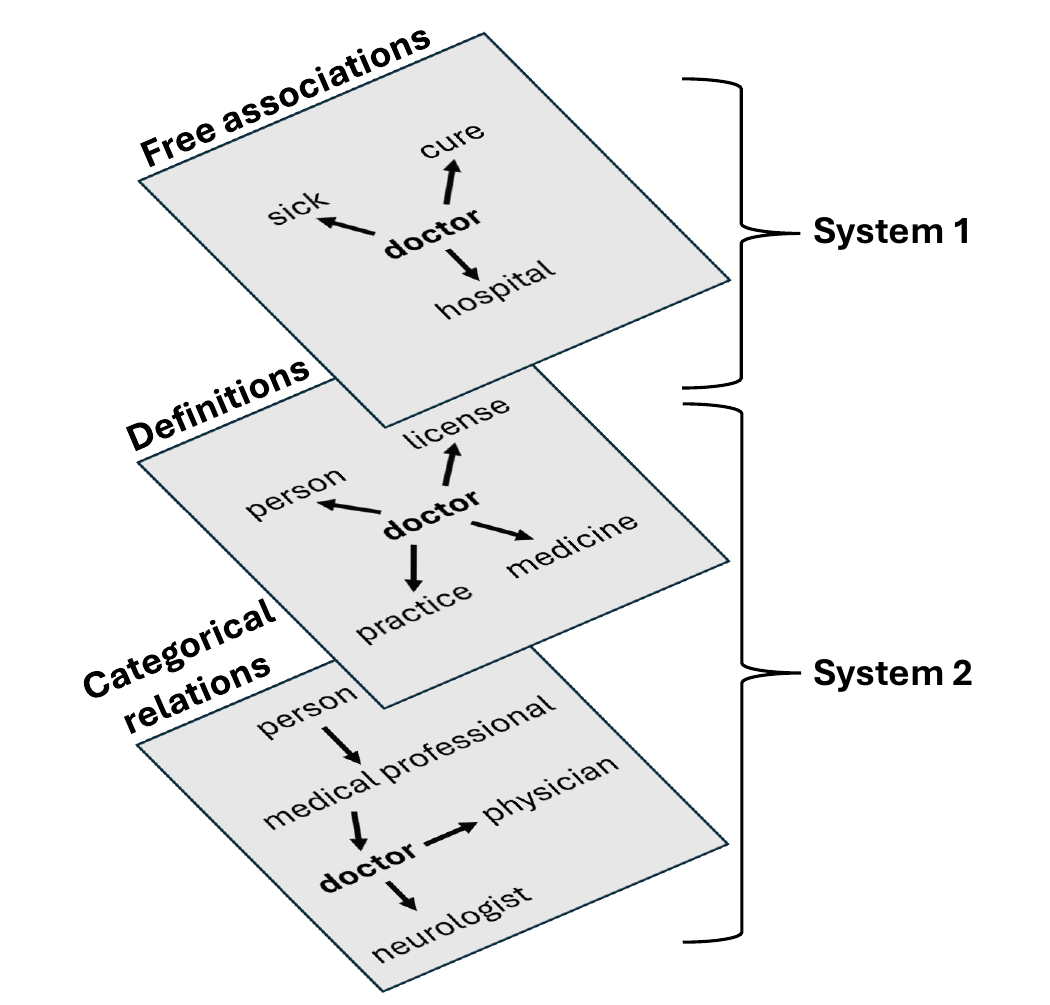}
\caption{\textbf{System~1 and System~2 semantic memory structures.} Three different structures of semantic memory are investigated, intended to approximate System~1 and System~2 thinking. System~1 semantic memory structures are built from free associations by connecting cue words to their associate responses. To approximate System~2 thinking, we build two different structures of semantic memory. First, we use definitions, connecting words to those words appearing in its definition, and then we use categorical relations, connecting words that share a categorical relationship such as synonyms, antonyms, hyponyms, and hypernyms.}
\label{fig:layers}
\end{figure}

We built three multilayer networks, one for humans, one for Mistral, and one for Llama3. Each multilayer network consists of three layers -- an associative layer built from free associations, a feature-based layer built from dictionary definitions, and a categorical layer built from WordNet relations (synonyms, antonyms, hyponyms, and hypernyms), as shown in Figure \ref{fig:layers}. The data used to build the multilayer networks came from diverse data sources, some that were pre-existing, and some that we created by prompting the LLMs, as described in what follows.

\smallskip
\noindent \textbf{Free Associations Layer}
For the Free Associations layers, we used the Small World of Words (SWOW) dataset \cite{de2019small} for the human network, and we used the LLM World of Words (LWOW) dataset \cite{abramski2025llm} for Mistral and Llama3 networks. The SWOW is a dataset of english free association norms, collected through a behavioral experiment in which thousands of human participants are given cue words (e.g. \textit{doctor}) and are asked to provide associative responses (e.g. \textit{sick, cure, hospital}) that come to mind. The dataset contains approximately 12,000 unique english cue words, and over 3 million associative responses. This set of approximately 12,000 cue words became the basis upon which other datasets were built, and we refer to it as the cue word set. The LWOW is similar to the SWOW, using the same set of cue words and following the same experimental procedure as the SWOW, but applied to LLMs (Mistral, Llama3, and Haiku). We used a modified version of SWOW dataset, which was preprocessed to align with the LWOW datasets generated by Mistral and Llama3. These datasets are available on the LWOW github page. From these three datasets (humans, Mistral, and Llama3), networks were built by connecting cue words to their responses, for example, by connecting \textit{doctor} to \textit{sick}, \textit{cure}, and \textit{hospital}. Edges were weighted based on response frequencies. The edges are inherently directed (from cue to response), but in order to facilitate analyses and to align with the other layers, we converted all edges to undirected edges, taking the larger of the two weights in cases of bidirectional edges. We then filtered these networks to decrease noise and reduce computational complexity, first by removing nodes not appearing in WordNet, then by removing idiosyncratic edges (weight = 1), and then finally by extracting the largest connected component of each network. The resulting networks are models of semantic memory that encode an implicit understanding of concepts, since the free associations are obtained without context and without reflection. Thus, this layer can be considered a proxy for System~1 thinking.

\smallskip
\noindent \textbf{Definitions Layer}
For the definitions layer, for humans, we used WordNet definitions of all words in the cue word set. For Mistral and Llama3, we generated WordNet definitions of words in the cue word set by prompting the models. First, for each word in the cue word set, we prompted the models to identify the various senses of each word (since several words can have more than one sense). Then, for each sense, we asked the models to provide definitions. In this way, the LLM-generated definitions closely resembled the WordNet definitions, in terms of structure, since WordNet also has different definitions for different word senses. Crucially, the word senses of each cue word identified by the LLMs do no necessarily match those in WordNet. From these datasets of definitions of words in the cue word set, networks were built by connecting each cue word to words appearing in its definition (or definitions if a cue word had more than one sense). For example, if we assume a doctor is defined as a person licensed to practice medicine, then \textit{doctor} would be connected to \textit{person}, \textit{licensed}, \textit{practice}, and \textit{medicine}. The edges were unweighted, and undirected, to facilitate further analyses. These networks were then filtered, similarly to the free association networks, first by removing nodes not appearing in WordNet, and then by extracting the largest connected component of each network. Networks built in this way capture the necessary and sufficient conditions of each concept, which is arguably a rather objective and logical way of defining a concept. Thus, this layer can be considered a proxy for System~2 thinking.

\smallskip
\noindent \textbf{Categorical Relations Layer}
For the categorical relations layer, for humans, we used WordNet relations (synonyms, antonyms, hyponyms, and hypernyms) of all words in the cue word set. For Mistral and Llama3, we generated WordNet relations of words in the cue word set by prompting the models. We used the same word senses that were identified by the models for generating the definitions. Therefore, for each word sense of each cue word, we asked the models to provide synonyms, antonyms, hyponyms, and hypernyms. In this way, the dataset closely resembled the WordNet relations. From these datasets, networks were built by connecting each cue word to all of its relations, for all senses of a word. For example, \textit{doctor} would be connected to its hypernym \textit{medical professional}, several hyponyms like \textit{neurologist}, \textit{cardiologist}, etc., and its synonym \textit{physician} (note that \textit{doctor} has no antonymn). The edges were neither weighted nor directed. These networks were then filtered, similarly to the other networks, first by removing nodes not appearing in WordNet, and then by extracting the largest connected component of each network. Networks built in this way capture the categorical meaning of a concept, which is a logical way of defining a concept, and can even be expressed mathematically. Thus, this layer can be considered a proxy for System~2 thinking.

By building multilayer networks in this way, we model cognition as a complex system with multiple structures, represented by the different layers. Each layer represents a different form of conceptual knowledge, which can be considered a proxy for either System~1 or System~2 thinking. This multilayer structure allows these layers to be compared empirically. Moreover, although the multilayer networks were constructed independently for humans and LLMs, they share the same underlying cue word set, ensuring that the resulting semantic structures are directly comparable across systems. In the following, we describe how this empirical comparison was carried out with structural reducibility analysis.

\subsection*{Structural reducibility analysis}

To assess whether the different semantic memory layers capture distinct forms of conceptual knowledge, we conducted a structural reducibility analysis of the multilayer networks \cite{de2015structural}. This method evaluates the extent to which layers in a multilayer network encode overlapping versus complementary structural information. Conceptually, the analysis determines whether multiple layers are necessary to preserve the organization of the system, or whether some layers can be merged without substantially altering the network’s structural properties. The method treats each layer as a network with its own adjacency matrix and compares layers using an information-theoretic distance measure. Specifically, each network layer is represented by a normalized Laplacian matrix, which captures the structural connectivity of the network. The similarity between layers is then quantified with a measure that provides an estimate of how much structural information differs between two layers. The algorithm then performs a hierarchical aggregation procedure in which layers that contain redundant information are iteratively merged. At each step of the procedure, the pair of layers whose aggregation results in the smallest loss of structural information is combined. After each aggregation step, the amount of information preserved in the reduced multilayer network is quantified using the relative entropy measure, which compares the structural information of the aggregated multilayer network to that of the fully separated multilayer representation.

The optimal multilayer configuration is identified as the point at which the relative entropy is maximized. Intuitively, this corresponds to the representation that preserves the greatest amount of structural information while minimizing redundancy between layers. If the optimal configuration occurs when all layers remain separate, the multilayer network is considered irreducible, indicating that each layer encodes distinct structural information. Conversely, if the optimal configuration merges some layers, this suggests that those layers contain overlapping or redundant structural patterns. Using the python package multired \cite{de2015structural}, we applied this analysis separately to the human and LLM multilayer networks comparing the three layers corresponding to associative, feature-based, and categorical knowledge. This allowed us to evaluate whether the conceptual knowledge structures represented by these layers form empirically distinct components of the semantic memory system.

\subsection*{Implicit bias measurement}
We next examined how different semantic memory structures relate to implicit bias within the networks. The goal of this analysis is to see if there are observable differences in measured implicit biases between System~1 and System~2 semantic memory structures.  We expected bias to be weaker in System 2 structures compared to System 1 structures, due to the type of conceptual knowledge encoded in these structures. Free associations tap into the implicit associative knowledge that is gained through years life experience (or training data in the case of LLMs), whereas the generation of definitions and categorical relations, in theory, taps into a more deliberative form of knowledge.

Free associations capture implicit associative knowledge acquired through experience (or training data in the case of LLMs), whereas definitions and categorical relations encode more structured and deliberative forms of conceptual knowledge. Stereotypes are often learned through real or perceived statistical regularities, resulting in associations between social identities and concepts (e.g. doctors tend to be male). However, more often than not, these social identities are completely independent from the definitional or taxonomic structure of the concepts they are associated with (e.g. the definition/category of doctor has nothing to do with being male). Thus, stereotypes should be less present in structured and deliberative representations. We therefore hypothesized that stereotype bias would be weaker in the System~2-like network layers than in the associative layer. In order to measure such bias, we used the bias evaluation methodology developed by Abramski et. al. (2025) \cite{abramski2025word}. Since we use a constant methodology for measuring bias, systematic differences in bias across layers would indicate that representational structure itself contributes to the differences in observed bias within the networks. For this analysis we focused on measuring gender stereotypes within the networks as a form of bias.

The bias evaluation methodology is conceptually related to the implicit association test (IAT), which measures the strength of associations between concepts by examining differences in response times during categorization tasks. Analogously, the present approach measures the strength of association between pairs of concepts within a semantic network representation.

The bias evaluation methodology follows the same idea as the implicit association test (IAT) for measuring bias in humans, which measures the strength of associations between pairs of concepts by examining differences in response times during categorization tasks. Analogously, the present approach measures the strength of association between pairs of concepts within a semantic network representation. Each pair consists of a prime and a target. Thus, to evaluate gender stereotype bias, we choose gender identities as primes (e.g. \textit{man, woman}) and gender stereotypical words as primes  (e.g. \textit{doctor, nurse}), and we measure the relative strength of association between stereotype consistent pairs like \textit{}{woman -- nurse} and \textit{man -- doctor} compared to stereotype-inconsistent pairs like \textit{man -- nurse} and \textit{woman -- doctor}. Stronger associations between stereotype-consistent pairs compared to their stereotype-inconsistent counterpart pairs are therefore indicative of gender bias.

We defined a set of primes corresponding to opposite gender identities, and a set of target words associated with common gender stereotypes. For the primes, we used five pairs of opposing gender identities, including the female identities \textit{woman, female, mother, girl, feminine}, and their male counterparts \textit{man, male, father, boy, masculine}. For the target words, we selected 172 words, half of which are related to female stereotypes and half of which are related to male stereotypes. These 172 words correspond to four different topics that are typically associated with gender stereotypes. The four topics include traits (e.g. \textit{emotional, aggressive}), home vs. career (e.g. \textit{laundry, business}), art vs. science (e.g. \textit{dance, math}), and professions (e.g. \textit{nurse, doctor}). The 172 targets divided among the four topics are shown in Table \ref{tab:stereotypes}. Combining each prime with each target, we obtain a total of 1,720 prime-target pairs. Each prime-target pair falls into one of the following categories: female-stereotype-consistent (\textit{woman – nurse}), female-stereotype-inconsistent (\textit{woman – doctor}) male-stereotype-consistent (\textit{male – doctor}) or male-stereotype-inconsistent (\textit{male – nurse}).

\begin{figure}[h]
\centering
\includegraphics[width=.7\linewidth]{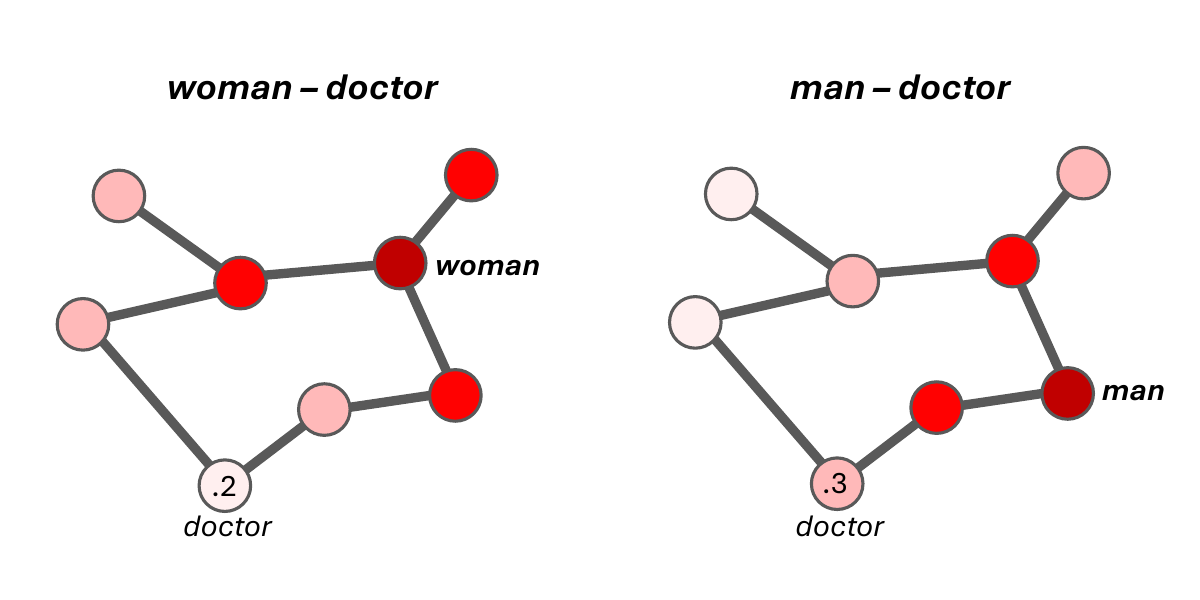}
\caption{\textbf{Measuring implicit bias with spreading activation.} The simplified example shows how the methodology can be used to evaluate gender bias by modeling semantic priming within a word association network. A spreading activation process is implemented in the network by activating the prime nodes \textit{woman} and \textit{man} independently. At the end of the process, the final activation level of \textit{doctor} is 0.2  when \textit{woman} is the prime node,  compared to 0.3 when \textit{man} is the prime node. This shows that the association between \textit{man} and \textit{doctor} is stronger than the association between \textit{woman} and \textit{doctor}, indicating gender bias. \cite{abramski2025word}}

\label{fig:spreading}
\end{figure}

The strength of association of each prime-target pair is measured by implementing spreading activation within each network layer of each multilayer network. This spreading activation is meant to simulate the cognitive phenomenon of semantic priming that occurs in humans. Semantic priming is the tendency for humans to recognize target words more quickly when prompted with a related prime word as opposed to an unrelated prime word, thus, reaction times can be used as a proxy to measure the strength of association between words \cite{hutchison2013semantic}. Previous work has shown that the semantic priming effect can be effectively simulated within semantic network models using spreading activation, thus enabling measures of strength of association between concepts within such network models \cite{siew2019spreadr,abramski2025llm}, and laying the groundwork for the methodology that we apply here \cite{abramski2025word}. In this spreading activation process, primes and targets correspond to nodes in the network. The process begins by activating the prime node within the network, representing exposure to a concept. The activation of that node then propagates along the edges of the network, spreading to other nodes through an iterative process that has a predetermined number of time steps. At the end of the process, the final activation of each node in the network represents the strength of association between that node and the initial activated prime node. Thus, by activating the prime nodes and observing the final activation levels of the target nodes at the end of the spreading activation process, we can obtain a measure of the strength of association between all the prime-target pairs. A simplified example of this process is captured in Figure \ref{fig:spreading} from Abramski et al. (2025) \cite{abramski2025word}. The idea is that if stereotype bias is strong within the network, the final activation levels of female-related targets will be greater when activated by female-related primes compared to male-related primes, and vice-versa. Instead, if stereotype bias is weak within the network, we will observe that the final activation levels of the targets do not change much when activated by  female-related primes compared to male-related primes.

\begin{table}[H]
\centering
\caption{\textbf{Target words.} The target words used for evaluating gender bias are shown in the tables. The targets, half of which are related to female stereotypes, and half of which are related to male stereotypes, fall into four categories: traits, home vs. career, art vs. science, and professions.}
\label{tab:stereotypes}

\setlength{\tabcolsep}{6pt}
\renewcommand{\arraystretch}{1.1}

\begin{tabular}{@{}c@{\hspace{4em}}c@{}}

\begin{tabular}{cc}
\hline
\multicolumn{2}{c}{\textbf{Traits}} \\ \hline
Female & Male \\ \hline
affectionate & active \\
cheerful & adventure \\
compassionate & aggression \\
connection & ambitious \\
consideration & assert \\
cooperative & athlete \\
depend & boast \\
emotional & competition \\
empathy & confidence \\
gentle & courageous \\
honest & determined \\
kind & dominant \\
loyalty & force \\
modest & hostility \\
nag & impulsive \\
nurture & independent \\
quiet & intellectual \\
sensitive & leader \\
support & logical \\
sympathetic & opinion \\
tender & outspoken \\
trusting & persistent \\
understanding & reckless \\
warm & stubborn \\
whine & superior \\ \hline
\end{tabular}
&
\begin{tabular}{cc}
\hline
\multicolumn{2}{c}{\textbf{Home/Career}} \\ \hline
Female & Male \\ \hline
home & career \\
house & work \\
household & job \\
kitchen & office \\
ironing & employee \\
cooking & supervisor \\
cleaning & business \\
bake & company \\
wash & professional \\
dishwasher & corporate \\
laundry & industry \\
sew & manager \\
shopping & hire \\
grocery & profession \\
decorate & ceo \\
family & promotion \\
parent & earn \\
spouse & money \\
child & invest \\
birth & success \\
infant & income \\
daycare & salary \\
marriage & paycheck \\
wedding & pension \\
caregiver & workplace \\ \hline
\end{tabular}

\\
\noalign{\vspace{4em}}

\begin{tabular}{cc}
\hline
\multicolumn{2}{c}{\textbf{Art/Science}} \\ \hline
Female & Male \\ \hline
art & math \\
poetry & science \\
dance & chemistry \\
pottery & physic \\
drawing & geology \\
painting & astronomy \\
sculpture & nasa \\
theater & engineering \\
cinema & computer \\
literature & technology \\
music & biology \\
photography & calculus \\
performance & equation \\
design & statistic \\
culture & mathematics \\
writing & economics \\
creativity & logic \\
expression & number \\ \hline
\end{tabular}
&
\begin{tabular}{cc}
\hline
\multicolumn{2}{c}{\textbf{Professions}} \\ \hline
Female & Male \\ \hline
artist & scientist \\
psychologist & carpenter \\
nurse & doctor \\
teacher & police officer \\
babysitter & firefighter \\
secretary & electrician \\
receptionist & plumber \\
therapist & engineer \\
maid & pilot \\
librarian & astronaut \\
dancer & mechanic \\
cashier & surgeon \\
designer & captain \\
technician & dentist \\
writer & mayor \\
singer & judge \\
clerk & contractor \\
journalist & farmer \\ \hline
\end{tabular}

\end{tabular}
\end{table}

Computationally, this spreading activation process is carried out using the R library spreadr \cite{siew2019spreadr}. This package requires the specification of a series of parameters that determine certain characteristics of the process, which we set following Abramski et al. (2025) \cite{abramski2025word}. The parameters include the initial activation level assigned to the prime node, which we set to the number of nodes in the network, the number of time steps, which we set to twice the diameter of the network, and the percentage of activation retained by the activated node, which we set to the default of 0.5.

Upon completion of the spreading activation process, the final activation levels are stored in an activation level matrix and normalized to control for frequency effects. For each stereotype topic, we then compute the difference in normalized activation levels between stereotype-consistent and stereotype-inconsistent prime–target pairs (e.g. the difference in normalized activation level between \textit{woman -- nurse} and \textit{man -- nurse}). These paired differences are aggregated to produce standardized effect sizes (using the Wilcoxon test for paired samples)  that quantify the magnitude and direction of bias within each network layer. Positive effect sizes indicate stereotype-consistent bias (e.g. stronger associations between male primes and male-related targets), whereas negative values indicate stereotype-inconsistent associations. Effect sizes near zero indicate little or no systematic bias. Effect sizes were computed separately for each network (humans, Mistral, and Llama3), for each layer (associative, definitional, categorical), and for each of the four stereotype topics.

\section{Results}

\subsection*{Multilayer network characteristics}

Basic statistics of the constructed multilayer networks are reported in Table \ref{tab:netstats}. We observe significant differences in the statistics across layers and across humans and LLMs. In humans, the categorical relations layers is largest in terms of nodes, but also the sparsest. The free associations layer is densest -- nearly six times as dense as the categorical relations layer, and almost twice as dense as the definitions layer, while the definitions layer is the smallest, with less than half the nodes of the categorical relations layer. In Mistral, similar to humans, the categorical relations layer is both the largest and the sparsest layer, but the layers are much more similar to each other in size and density compared to humans. The categorical layer in Mistral is also much denser compared to the same layer in humans. In Llama3, in contrast to humans and Mistral, the free associations layer is the largest of the layers, and in fact, it is the largest of all the layers of all the networks, but slightly less dense compared to the same layer in humans and Mistral. The definitions layer in Llama3 is the smallest in terms of nodes, like in humans and Mistral, but unlike in humans, it is also the densest layer. The categorical relations layer is very sparse, similar to humans. Notably, in humans and Llama3, the size and density of the network layers are drastically different between layers, while in Mistral, the layers are more similar to each other. However, in terms of comparisons across networks, Mistral is more similar to humans in that the categorical relations layer is the largest, followed by free associations and then definitions.  Although the overall sizes of the networks differ somewhat across humans, Mistral, 
and Llama3, the three-layer structure is broadly comparable across systems. This 
comparability allows the multilayer structures to be analyzed using the same reducibility and bias evaluation procedures.

\begin{table}[H]
\centering
\caption{\textbf{Network statistics.} The table shows network statistics by network and layer, including number of nodes, number of edges, network density, and mean degree.}
\begin{tabular}{llcccc}
\hline
\textbf{Network} & \textbf{Layer}       & \textbf{Nodes} & \textbf{Edges} & \textbf{Density} & \textbf{Mean degree} \\ \hline
\textbf{}        & Free Associations    & 24,308         & 317,344        & 0.0011           & 26                      \\
Humans           & Definitions          & 16,200         & 83,971         & 0.0006           & 10                      \\
\textbf{}        & Categorical relations & 34,211         & 94,541         & 0.0002           & 6                       \\ \hline
\textbf{}        & Free Associations    & 20,339         & 199,103        & 0.0010           & 20                      \\
Mistral          & Definitions          & 16,806         & 153,771        & 0.0011           & 18                      \\
\textbf{}        & Categorical relations & 26,451         & 236,931        & 0.0007           & 18                      \\ \hline
\textbf{}        & Free Associations    & 38,987         & 546,866        & 0.0007           & 28                      \\
Llama3           & Definitions          & 18,526         & 214,172        & 0.0012           & 23                      \\
\textbf{}        & Categorical relations & 27,676         & 110,430        & 0.0003           & 8                       \\ \hline
\end{tabular}
\label{tab:netstats}
\end{table}

\subsection*{Structural reducibility differs between humans and LLMs}

\begin{figure}
\centering
\includegraphics[width=1\linewidth]{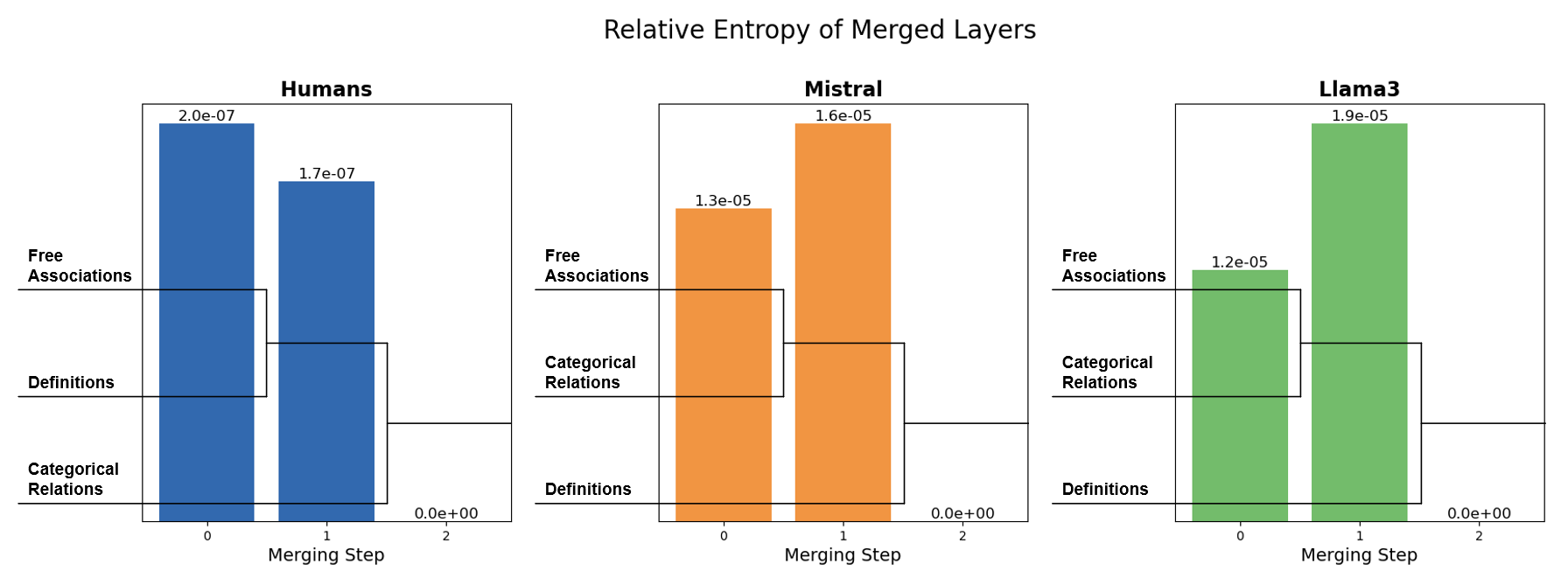}
\caption{\textbf{Structural reducibility of multilayer semantic networks.}
Relative entropy values are shown for different layer aggregation configurations. For humans (blue), entropy is maximized when all three layers remain separate.
For Mistral (orange) and Llama3 (green), entropy is maximized when the associative and categorical layers are merged, indicating partial reducibility.}
\label{fig:reducibility}
\end{figure}

The results of the structural reducibility analysis are shown in Figure 
\ref{fig:reducibility}. For the human multilayer network, the relative entropy 
measure was maximized when all three layers remained separate, indicating that 
the multilayer structure is irreducible. Merging any pair of layers resulted in 
a measurable loss of structural information, suggesting that the associative, 
definitional, and categorical relations each contribute distinct connectivity 
patterns to the organization of semantic memory. These results indicate that human semantic memory is characterized by multiple irreducible representational structures, consistent with the idea that different forms of conceptual knowledge are maintained in differentiated memory systems. In contrast, the reducibility analysis revealed a different pattern for the LLM networks. For both Mistral and Llama3, the optimal configuration involved a partial collapse of layers. Specifically, the associative and categorical layers could be merged with minimal loss of structural information, while the definitional layer remained relatively distinct. This suggests that, in these models, different types of semantic relations are less differentiated at the representational level than in humans. Although the networks contain information corresponding to associations and category structure, these forms of knowledge appear to be encoded in a more overlapping manner. 

The results of the structural reducibility analysis are shown in Figure \ref{fig:reducibility}. For the human semantic networks, the analysis indicated that the optimal representation retained all three layers as distinct. Merging any pair of layers resulted in a measurable loss of structural information, suggesting that each layer contributes uniquely to the overall organization of semantic memory. In particular, associative, definitional, and categorical relations exhibited different patterns of connectivity and organization that could not be reduced to one another without compromising the integrity of the network.

\noindent Taken together, these results indicate that the semantic memory layers are 
structurally distinct in humans but partially redundant in the LLM networks. 
In particular, associative and categorical relations show substantial overlap 
in the LLM representations, whereas they remain structurally differentiated in 
the human network.


\subsection*{Relationship between semantic structure and bias differs across humans and LLMs}

\begin{figure}
\centering
\includegraphics[width=.9\linewidth]{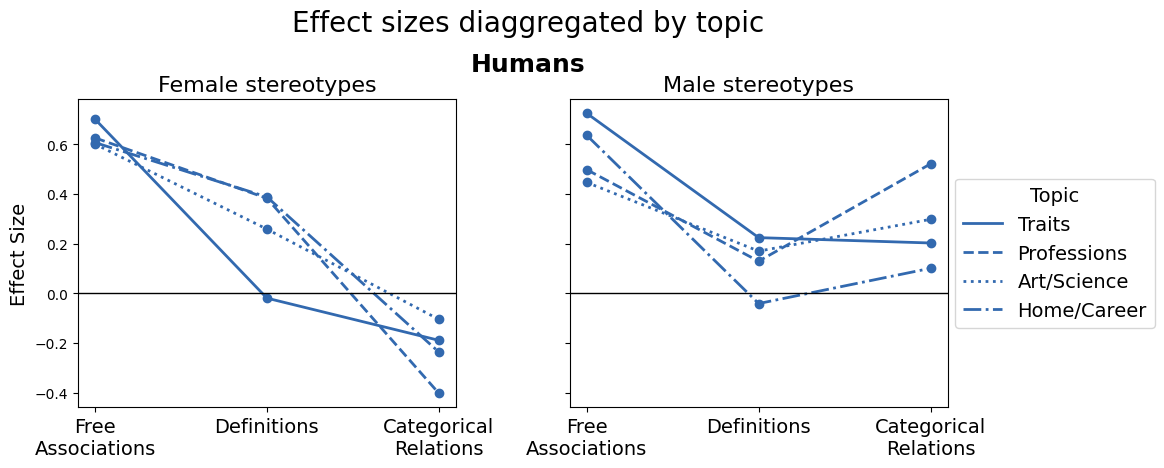}
\includegraphics[width=.9\linewidth]{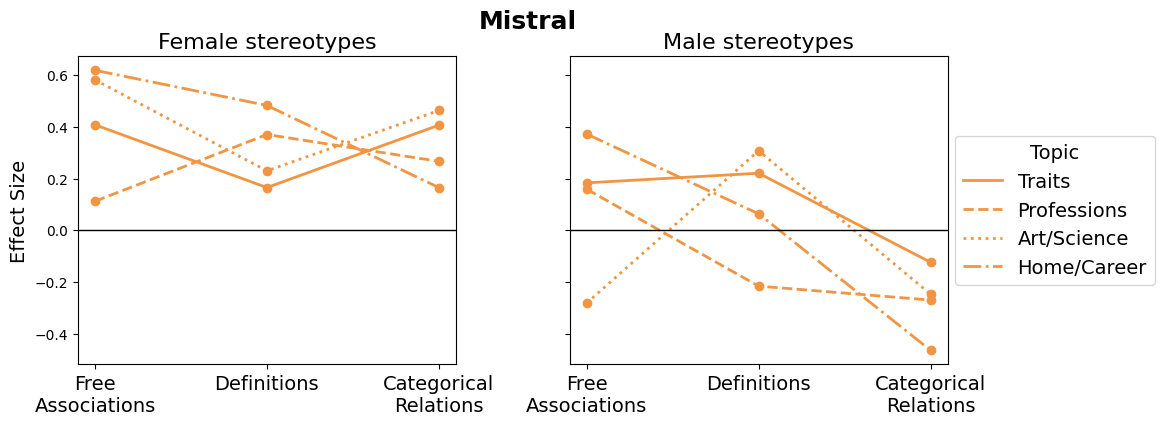}
\includegraphics[width=.9\linewidth]{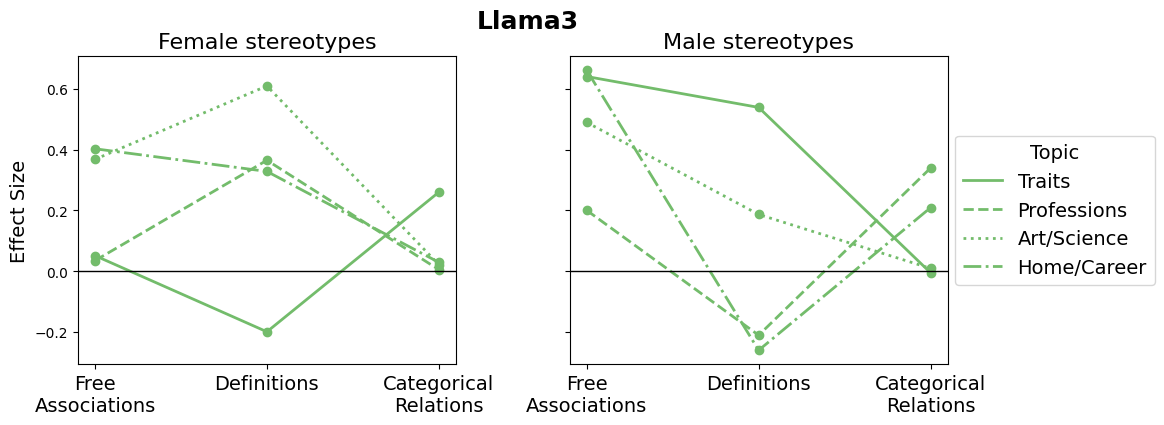}
\caption{\textbf{Effect sizes disaggregated by stereotype topic.}
Effect sizes measuring stereotype-consistent associations are shown separately for humans, Mistral, and Llama3 across four stereotype topics: traits, professions, home/career, and arts/sciences. Within each network, effect sizes are computed for the associative, definitional, and categorical layers. Positive values indicate stereotype-consistent associations, while negative values indicate stereotype-inconsistent associations. The plots illustrate how bias varies across semantic structures and topics within each network.}
\label{fig:effects_disagg}
\end{figure}

The spreading activation analysis produced activation level matrices for each network and layer, from which effect sizes were computed for all prime–target pairs. Effect sizes were calculated separately for female-related targets and male-related targets, for each network (humans, 
Mistral, Llama3), each network layer (free associations, definitions, categorical relations), and each stereotype topic (traits, professions, home/career, and arts/sciences). The resulting effect sizes are shown in Figure \ref{fig:effects_disagg}. In addition to the effect sizes, the normalized activation level matrices are represented as heatmaps in a series of figures in Appendix \ref{appendixA}. From these heatmaps, it is possible to see the relative strength of association between primes and targets, painting a qualitative picture of the types of gender biases exhibited by humans, Mistral, and Llama3, that go beyond the qualitative effect sizes.

\noindent From these results, clear differences between humans and LLMs emerged. The most notable difference is that bias patterns are consistent across all four topics in humans but not in LLMs. This is clearly visible in the plots in Figure \ref{fig:effects_disagg}. The blue lines (humans) corresponding to different stereotype topics all follow similar trajectories, while the orange and green lines (LLMs) follow different trajectories depending on the stereotype topic. Specifically, in the human network across all four topics, the associative layer consistently exhibited the strongest stereotype-consistent effects, while both System~2-like layers -- the definitions and categorical relations layers -- showed substantially reduced  bias relative to the associative layer. The fact that this tendency is stable across topics indicates a consistent pattern in which bias decreases as representations shift from associative links to more structured semantic relations. Instead, in LLMs, such a consistent patten across stereotype topics is lacking. Within the same layer, effect sizes are drastically different for different topics. For example, in Llama3, in the definitions layer, male-stereotype-consistent effects are observed for the traits and art vs. science, but male-stereotype-inconsistent effects are observed for professions and home vs. career. This inconsistency suggests that stereotype expression in LLMs is random and unstable, while in humans it depends on distinct underlying conceptual representations. We also observe that differences in effect sizes between layers are less drastic, especially for female-stereotypes in Mistral. This further implies that there is no clear relationship between bias and semantic memory structure in LLMs.

An additional pattern emerged in the categorical relations layer in humans. For female-related targets, effect sizes were negative, indicating stereotype-inconsistent associations, but for male-related targets, effect sizes remained positive. This suggests that both female-related targets like \textit{nurse} and \textit{laundry} as well as male-related targets like \textit{doctor} and \textit{business} are more strongly associated with the male gender. This asymmetry suggests that taxonomic knowledge in humans may be dominated by male conceptual categories, counteracting female stereotypes but over-compensating in the other direction.

\begin{figure}[H]
\centering
\includegraphics[width=1\linewidth]{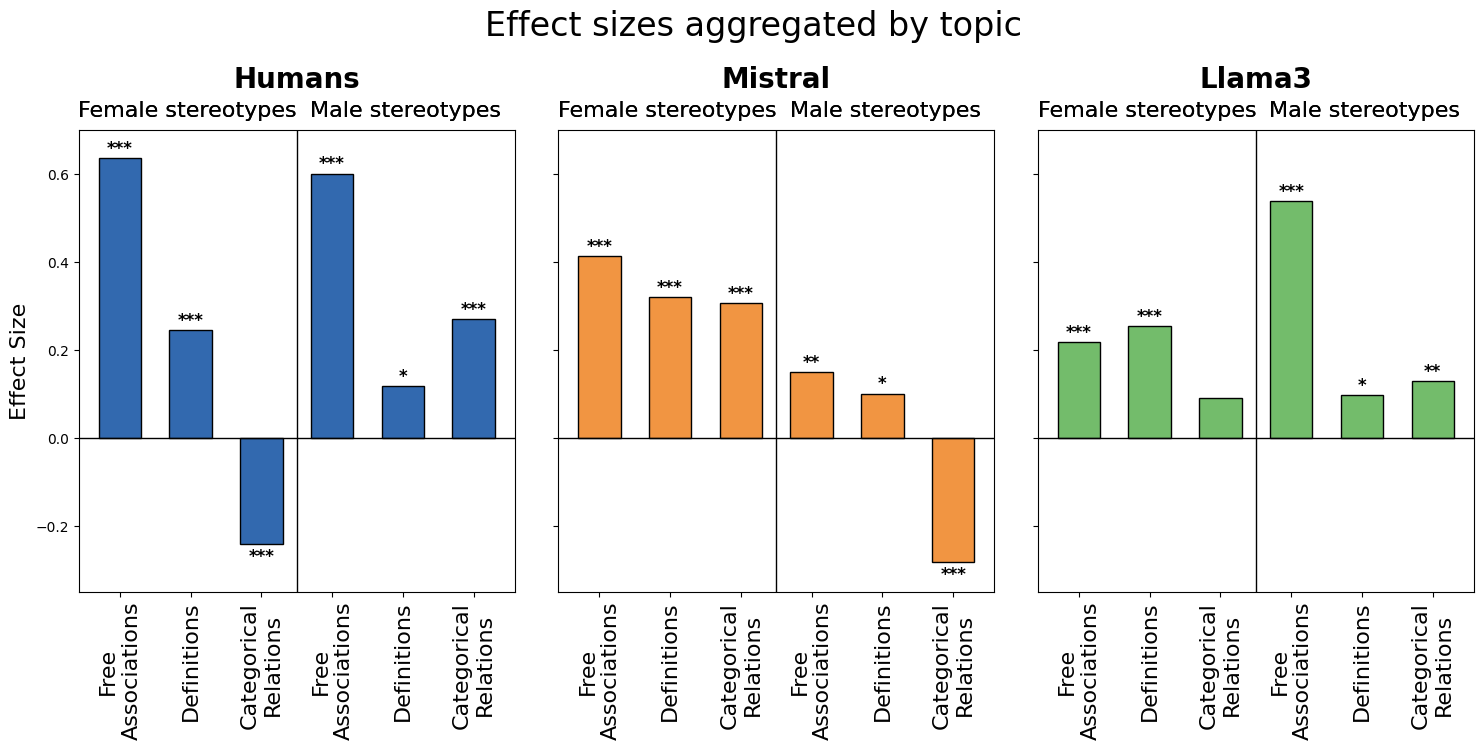}
\caption{\textbf{Effect sizes aggregated by stereotype topics.} Effect sizes are aggregated across the four stereotype topics to provide an overall measure of stereotype bias within each network and semantic layer. Positive values indicate stereotype-consistent associations, whereas negative values indicate stereotype-inconsistent associations. The results summarize how bias varies across associative, definitional, and categorical representations in humans, Mistral, and Llama3.}
\label{fig:effects_agg}
\end{figure}

To assess overall trends, effect sizes were aggregated across the four stereotype topics. The aggregated results are shown in Figure \ref{fig:effects_agg}. From these plots, we observe the overall relationship between bias and semantic memory structure more clearly. In the human network, bias decreases systematically from the associative layer to the definitional and categorical layers. In contrast, Mistral and Llama3 show no consistent structural pattern: bias levels vary across layers without the systematic reduction observed in human semantic memory, indicating that semantic memory structure does not reliably predict bias magnitude in LLMs.

Together, these results reveal a clear difference between humans and LLMs. In humans, semantic memory structure and implicit bias vary systematically across layers of conceptual knowledge, regardless of stereotype topic, while in LLMs, the relationship between bias and semantic memory structure is inconsistent across stereotype topics and shows no clear pattern.


\section{Discussion}

The present study examined how System~1 and System~2 semantic memory structures relate to implicit bias in humans and LLMs by representing different forms of conceptual knowledge as multilayer semantic networks. Two main findings emerged. First, the structural reducibility analysis showed that the semantic memory layers are irreducible in humans but only partially reducible in the LLMs. Second, semantic memory structure was consistently related to implicit bias in humans, but not in the LLMs. Taken together, these results suggest that the organization of conceptual knowledge differs in important ways between humans and the studied language models, and that these differences affect how bias is expressed within their semantic representations.

\subsection*{Irreducible conceptual structures in human semantic memory}

The reducibility analysis revealed a clear structural difference between humans and the LLMs. In the human network, the free associations, definitions, and categorical relations layers were irreducible: merging any pair of layers resulted in a measurable loss of structural information. This indicates that these three forms of conceptual knowledge are organized as distinct structures within human semantic memory. In contrast, both Mistral and Llama3 showed partial reducibility. In these networks the free associations and categorical relations layers could be merged with little loss of structural information, while the definitions layer remained more distinct. One possible explanation for this pattern is that human conceptual knowledge includes different types of semantic relations that are not easily collapsed into a single structure. Associative relations reflect statistical regularities and contextual experience, definitions capture feature-based conceptual descriptions, and categorical relations encode logical or hierarchical relations between concepts. The fact that these layers are irreducible in the human network suggests that these forms of knowledge are maintained as separate structures rather than simply different expressions of the same underlying representation.

LLMs on the other hand appear to lack the capacity to produce certain categorical relations that are significantly distinct from free associations. In fact, upon closer examination of the categorical relations datasets, we found that the LLMs frequently produced relations that were associative rather than truly categorical. This was especially true for cases in which a true categorical relation is nonsensical, for example, the antonym of a concept like \textit{chair}. A human knows that such a concept does not have an antonym, but we found that the LLMs were likely to provide associative responses like \textit{table} rather than providing nothing at all. This tendency is reflected in the network statistics: the categorical relations layer is extremely sparse in humans, while it is comparatively much less sparse in the LLMs. This tendency also helps explain why the associative and categorical layers collapse in the reducibility analysis. If the categorical relations generated by the models are often associative in nature, then the two layers will naturally overlap. Humans, in contrast, appear to maintain clearer distinctions between associative, definitional, and categorical knowledge.

These results have implications for both cognitive science and AI. From the perspective of cognitive science, the findings provide network-level evidence that human semantic memory contains multiple structurally distinct forms of conceptual knowledge. The irreducibility of the layers suggests that associative, definitional, and categorical relations are not simply different ways of describing the same structure, but instead represent different components of semantic organization. For LLMs, the results suggest that some LLMs may not capture the full range of relational structure present in human conceptual knowledge. In particular, categorical knowledge appears to be weakly represented and often replaced by associative structure. This may help explain why language models sometimes struggle with tasks that require clear hierarchical reasoning or precise semantic distinctions.

\subsection*{Semantic memory structure and the regulation of bias}

The second main finding concerns the relationship between semantic memory structure and implicit bias. In the human multilayer network, we observed a clear and consistent pattern of bias across the four stereotype topics. The associative layer consistently exhibited the strongest stereotype-consistent effects. In contrast, both System~2-like layers -- the definitions and categorical relations layers -- showed substantially reduced bias, confirming our initial hypothesis. These results are consistent with the nature of stereotypes, which are learned and reinforced through exposure to associations. Evidently, these associations were captured by the free associations datasets used in this study, and subsequently encoded in the semantic memory networks. Definitions and categorical relations on the other hand encode different types of knowledge that are largely independent from associations. Definitions focus on the necessary or descriptive features of a concept, while categorical relations describe logical or taxonomic aspects of a concept. Thus, these forms of representation are less directly tied to stereotypical associations, which may explain why bias appears weaker in these layers.

In contrast to humans, we observed that the LLM networks did not show the same consistent pattern across topics. Effect sizes differed drastically within the same layer for different topics, and at the same time, differences in aggregated effects between layers were less pronounced compared to humans. In other words, the relationship between semantic structure and bias proved to be inconsistent, unpredictable, and unstable. This lack of consistency may be related to the structural overlap observed in the reducibility analysis. If associative and categorical knowledge are encoded in largely overlapping ways, then activating different semantic structures may not meaningfully change the pattern of associations in the network. In humans, different types of conceptual knowledge appear to provide distinct representational pathways. In the LLMs, those pathways may be less differentiated. Additionally, there may be some latent factors at play that cause such drastic differences between topics within the same layer. For example, in Llama3 in the definitions layer, for female stereotypes, we observe stereotype-consistent effects for professions but stereotype-inconsistent effects for traits, while the exact opposite pattern is observed for male stereotypes. This means that in definition based semantic representations produced by Llama3, all professions are more strongly associated to the male gender, while all traits are more strongly associated to the female gender (on average). This suggests that the stereotypes are more strongly tied to the topic categories themselves rather than to the individual concepts. This does not imply that LLMs are less biased than humans, rather, it suggests that the semantic representations are of a completely different nature compared to those of humans. Such differences are likely due to how the semantic representations are learned from the training data.

Another notable result is the presence of opposite effect size magnitudes in the human categorical relations layer. For female-related targets, the effect size was negative, indicating stereotype-inconsistent associations, while for male-related targets, the effects were weakly stereotype-consistent. This suggests that the male gender is more strongly associated with all targets (on average). One possible explanation is that categorical relations in WordNet are dominated by male defaults. For example, the concept \textit{man} in WordNet may include the sense that refers to \textit{mankind}. This reflects the historical use of the word \textit{man} to describe all people, both male and female, such as the idea that dogs are "man's best friend". This seems like a viable explanation, since many professions and socially valued roles have long been associated with men, so these historical patterns are likely embedded within the categorical hierarchies included in WordNet. As a result, categorical relations may weaken some female-stereotype-consistent associations while still preserving residual male biases within the hierarchy.

Overall, these results suggest that bias regulation in humans may depend not only on deliberate reasoning but also on the structure of semantic memory itself. When concepts can be accessed through different representational structures, those structures may differ in how strongly they express stereotypical associations. In the LLMs studied here, this structural mechanism appears to be largely absent.

\subsection*{Limitations}

The results of our analyses are grounded in well-studied theories and methodologies from network science and cognitive science. Nevertheless, several limitations should be considered when interpreting these findings. First, the human datasets used in this study come from English-speaking participants and therefore reflect cultural patterns specific to Western societies. Semantic structures and stereotype patterns may differ in other cultural or linguistic contexts. Future work should examine whether similar patterns emerge in semantic networks derived from other languages and populations. Second, the primes and targets used in the analyses are limited to the words that appear in the constructed networks. Although the networks are large, they represent only a subset of possible concepts that may be better suited for investigating stereotypes. For example, the concept \textit{assertive} was not present in all networks, and so we instead used the concept \textit{assert}, however, such a difference may influence in the results due to the difference in grammatical category. Third, the LLM networks were constructed using datasets generated through prompting the LLMs. While this approach makes it possible to construct comparable datasets across humans and models, LLMs have been shown to be highly sensitive to variations in prompts. Therefore, the results may depend partly on how the prompts were formulated. Different prompts or generation parameters could lead to somewhat different relational structures. Finally, as we discussed earlier in this manuscript, one of the main advantages of this approach is that it focuses explicitly on representational structure, enabling empirical investigations of the two distinct memory systems (associative and rule-based). This advantage, however, can also be considered a limitation because it abstracts away from the rule-based operations that characterize System~2-like reasoning. Therefore, it is important to note that we use this multilayer network approach to approximate System~1 and System~2 reasoning, recognizing that the approach fails to take into account many aspects of the dual process theory that may be captured by other more theoretical approaches.

\subsection*{Conclusion}

In this study, we investigated the relationship between System~1 and System~2 semantic memory structure and bias in both humans and LLMs using multilayer semantic memory networks. We found that human semantic memory contains structurally distinct forms of conceptual knowledge, while LLMs may lack some forms of conceptual knowledge possessed by humans, specifically, relational knowledge. We also found that semantic memory structure relates consistently to gender bias in humans, with lower levels of bias in System~2 memory structures. On the contrary, such a consistent pattern was not observed in Mistral and Llama3, suggesting that expressions of stereotype bias are regulated by drastically different semantic mechanisms compared to humans. These findings highlight important differences between human and machine semantic representations. Moreover, they suggest that the structure of conceptual knowledge should be taken into consideration when thinking about how bias emerges and is regulated in both humans and LLMs.

\section*{Statements and Declarations}

\subsection*{Availability of data and materials}
\noindent All data and code will be made available upon request.

\subsection*{Competing Interests}
\noindent The authors declare no competing interests.

\subsection*{Funding}
\noindent This work is supported by: (i) SoBigData.it which receives funding from the European Union -- NextGenerationEU -- National Recovery and Resilience Plan (PNRR) of Italy; (ii) the Future Artificial Intelligence Research (FAIR) Foundation, a non-profit foundation financed under the National Recovery and Resilience Plan (PNRR) of Italy; (iii) UniTrento Internal Call for Research 2023 grant from the Università degli Studi di Trento (Grant ID: PS 22\_27). 

\subsection*{Author contribution}
\noindent K.A., G.R., and M.S. conceived the methodology. K.A. conducted the analyses and analyzed the results. K.A., G.R., and M.S. interpreted the results and reviewed the manuscript.

\vspace{6pt}

\appendix
\section[\appendixname~\thesection]{Heatmaps representing normalized activation level matrices}
\label{appendixA}

\begin{figure}[H]
\centering
\includegraphics[width=.9\linewidth]{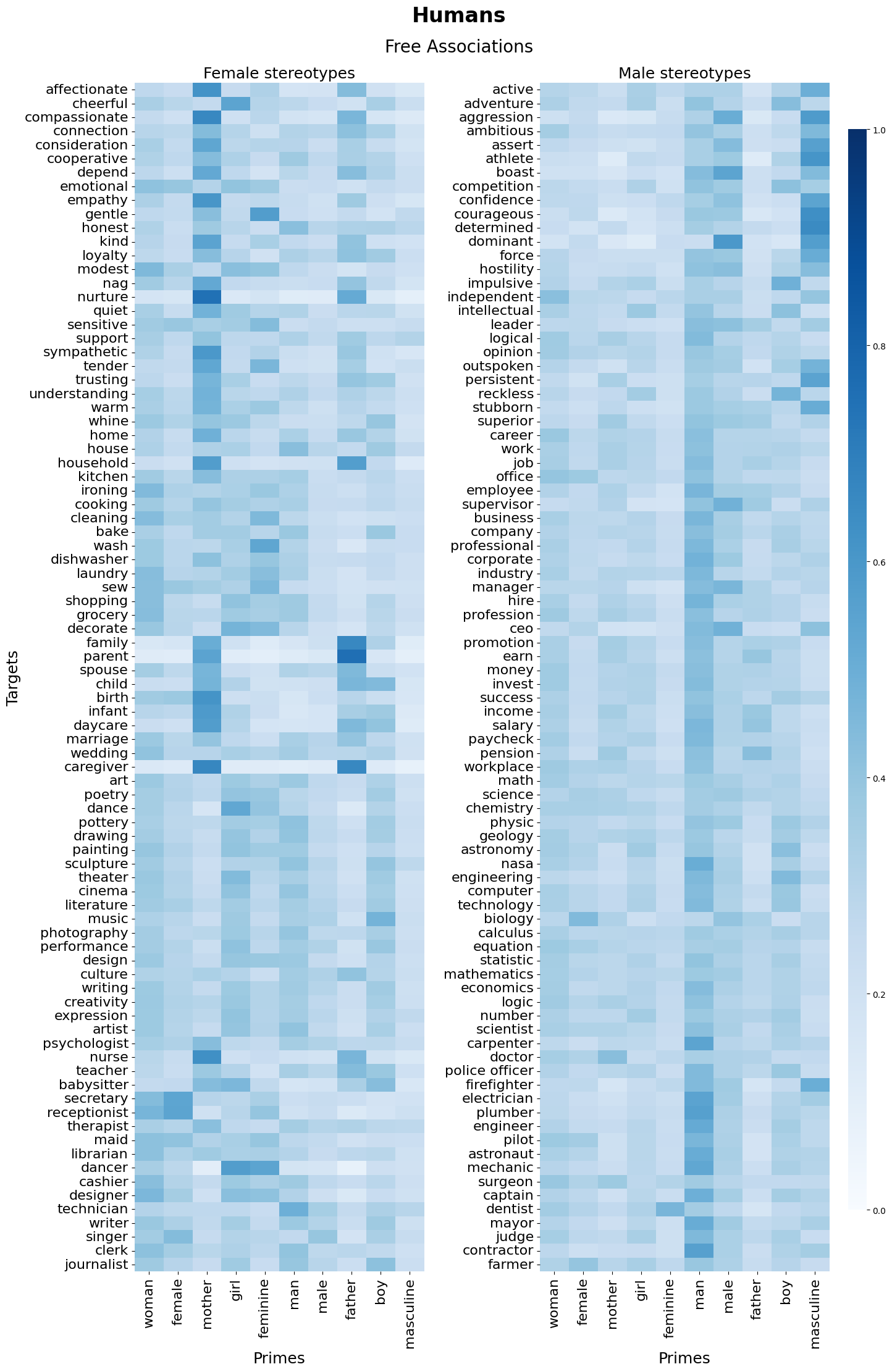}
\caption{\textbf{Normalized activation levels for the free associations layer in humans.} Strong associations are observed between certain prime-target pairs like \textit{mother -- nurture} and \textit{father -- parent}. The male-stereotypes heatmap shows that activation levels are notably greater for male-related targets when activated by male-related primes, evidenced by overall darker shades on the right side of the plot.}
\end{figure}

\begin{figure}[H]
\includegraphics[width=.9\linewidth]{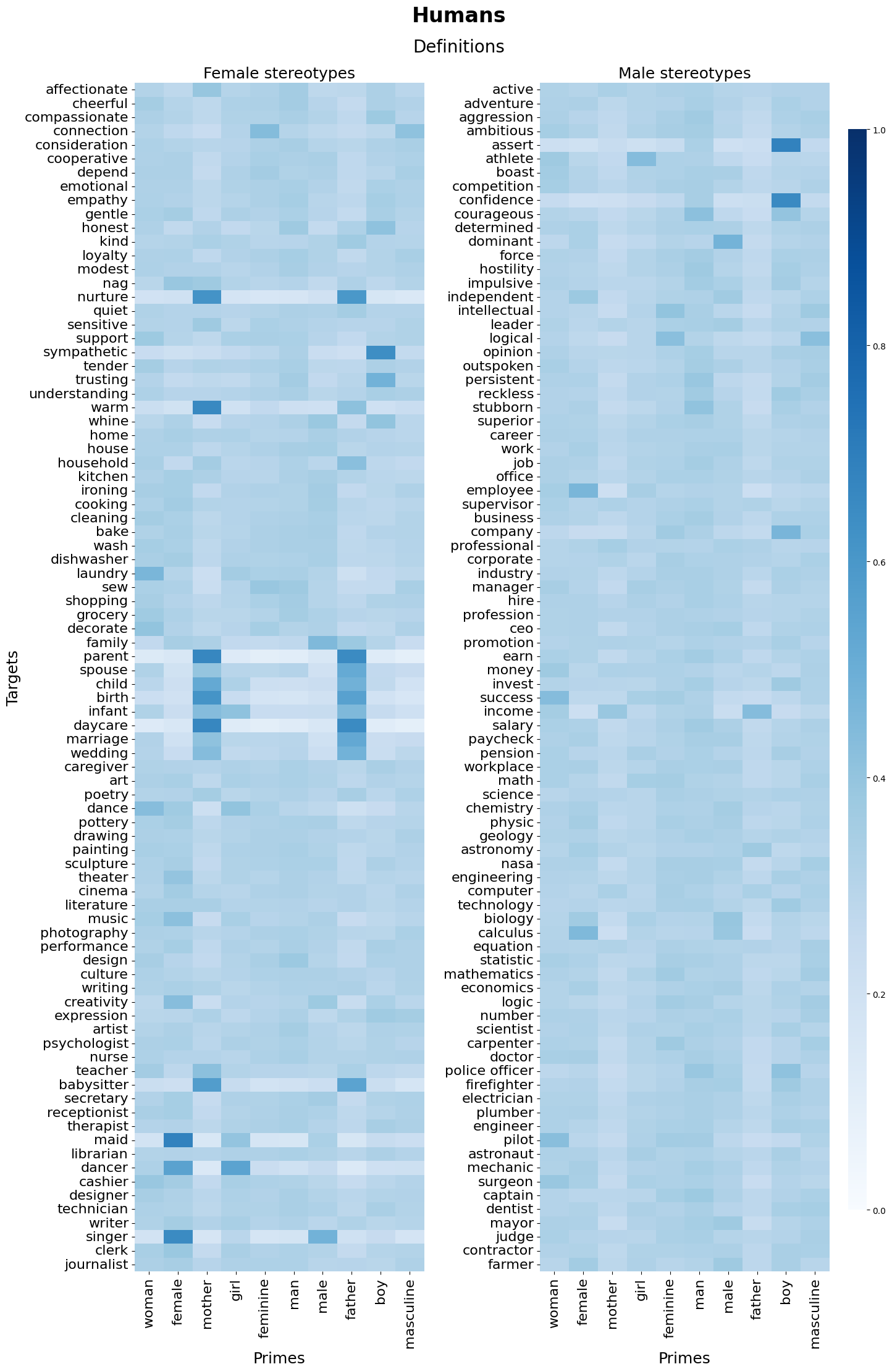}
\caption{\textbf{Normalized activation levels for the definitions layer in humans.} A large degree of symmetry is observed in the heatmaps, such as the concepts \textit{parent}, \textit{daycare}, and \textit{birth} being activated similarly by \textit{mother} and \textit{father}. There is little visible evidence of differences in activation levels between stereotype-consistent prime-target pairs and stereotype-inconsistent prime-target pairs, reflecting low levels of bias in this layer.}
\end{figure}

\begin{figure}[H]
\includegraphics[width=.8\linewidth]{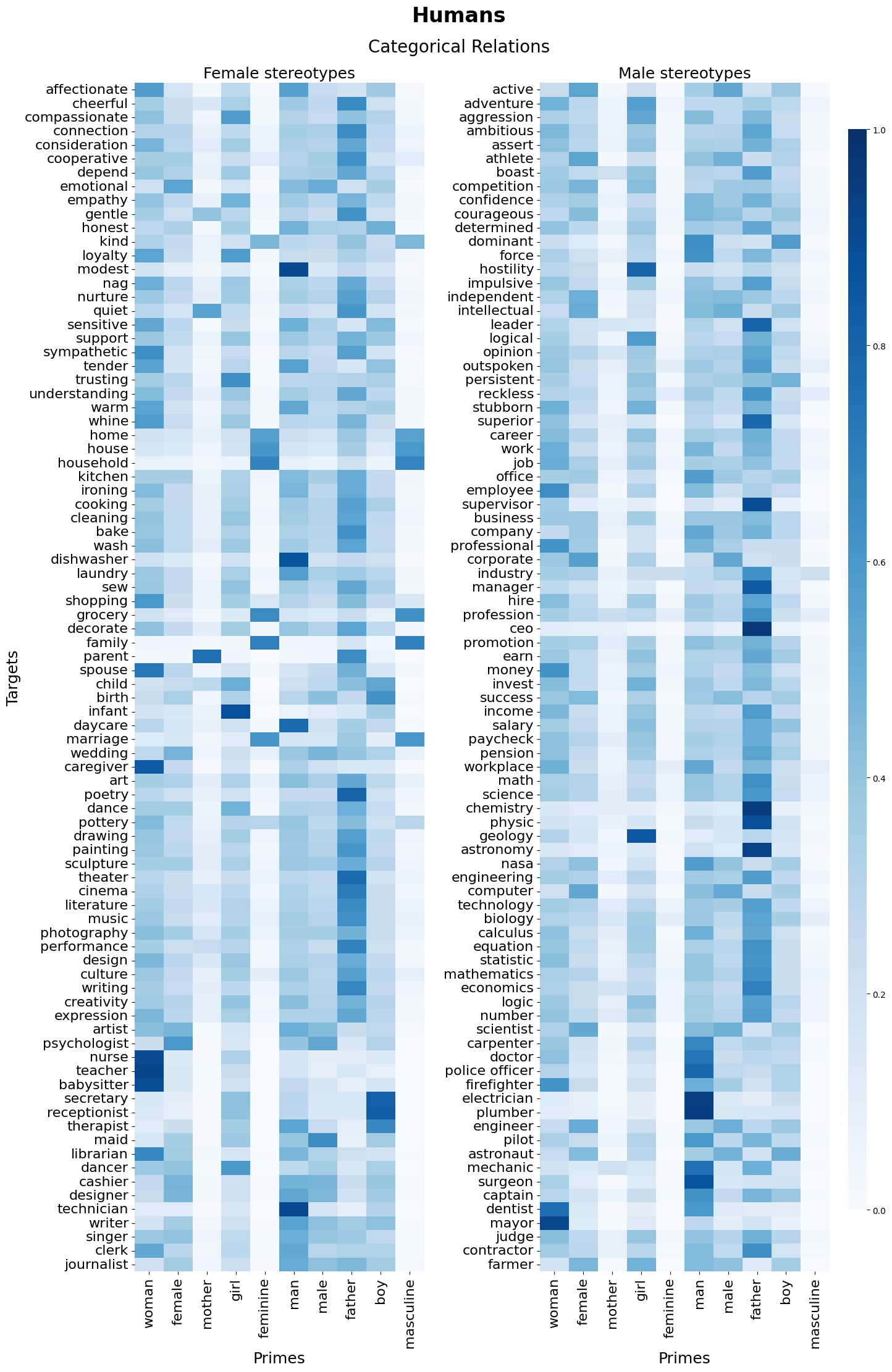}
\caption{\textbf{Normalized activation levels for the categorical relations layer in humans.} The heatmaps show large variations in the normalized activation levels of prime-target pairs, in contrast to the free associations and definitions layers.}
\end{figure}

\begin{figure}[H]
\includegraphics[width=.8\linewidth]{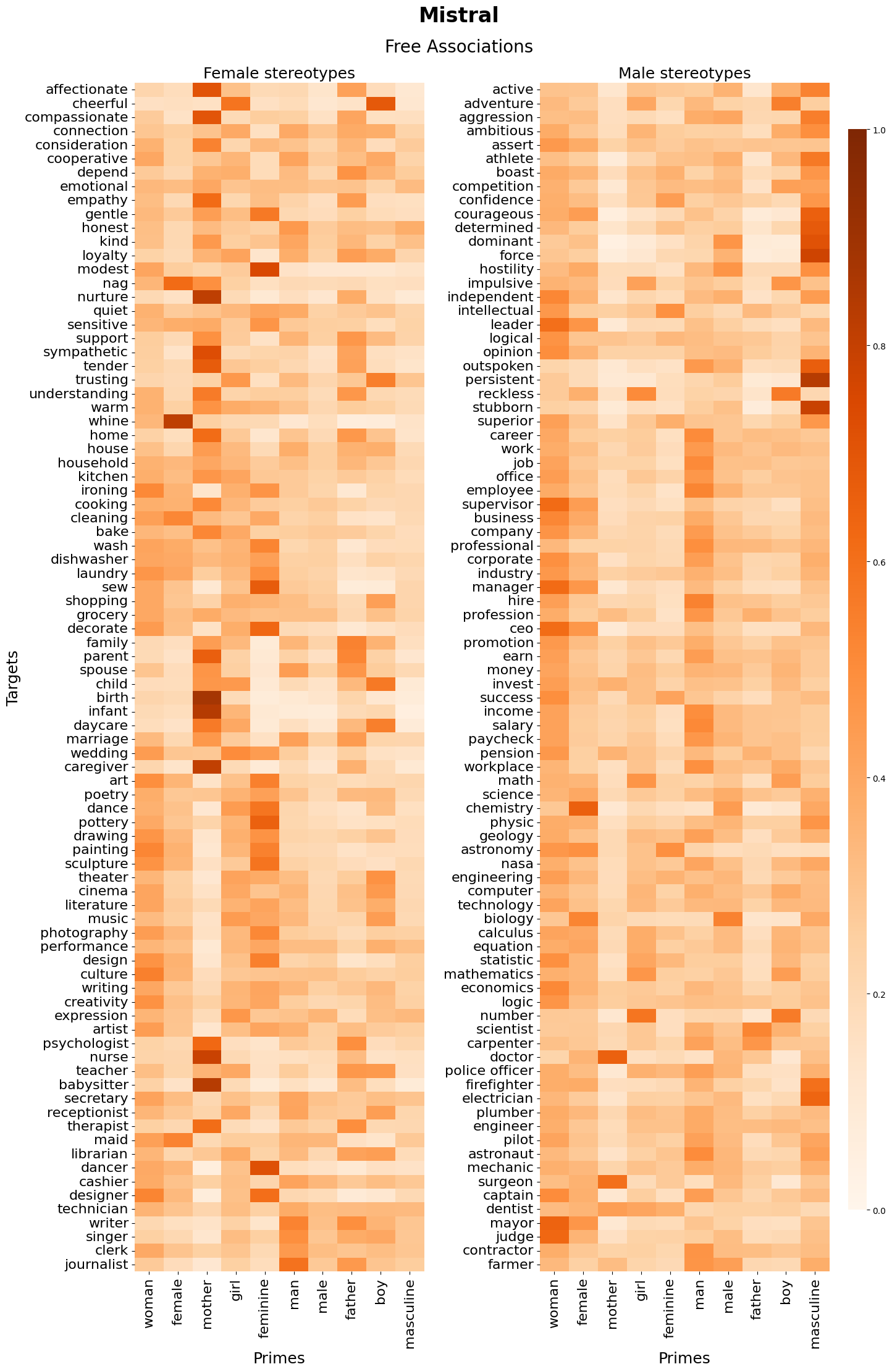}
\caption{\textbf{Normalized activation levels for the free associations layer in Mistral.} Strong associations are observed between certain prime-target pairs like \textit{mother -- birth} and \textit{masculine -- persistent}. It appears that certain stereotype topics are particularly activated by certain primes (male-related traits are strongly activated by the prime \textit{masculine}.}
\end{figure}

\begin{figure}[H]
\includegraphics[width=.8\linewidth]{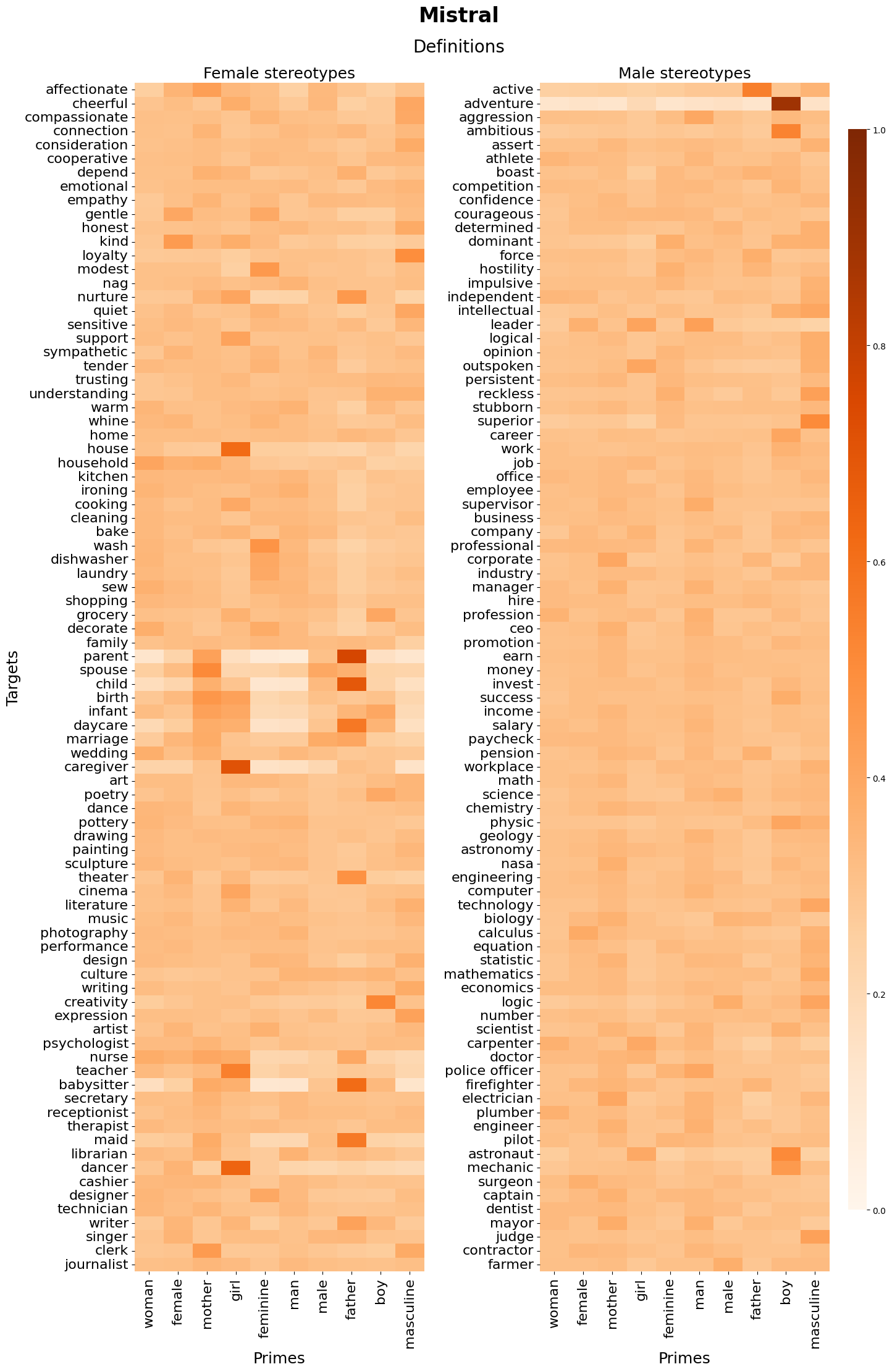}
\caption{\textbf{Normalized activation levels for the definitions layer in Mistral.} Normalized activation levels appear mostly homogeneous, with little variation across topics and layers, with the exception of a few prime-target pairs like \textit{boy -- adventure} and \textit{father -- parent}.}
\end{figure}

\begin{figure}[H]
\includegraphics[width=.8\linewidth]{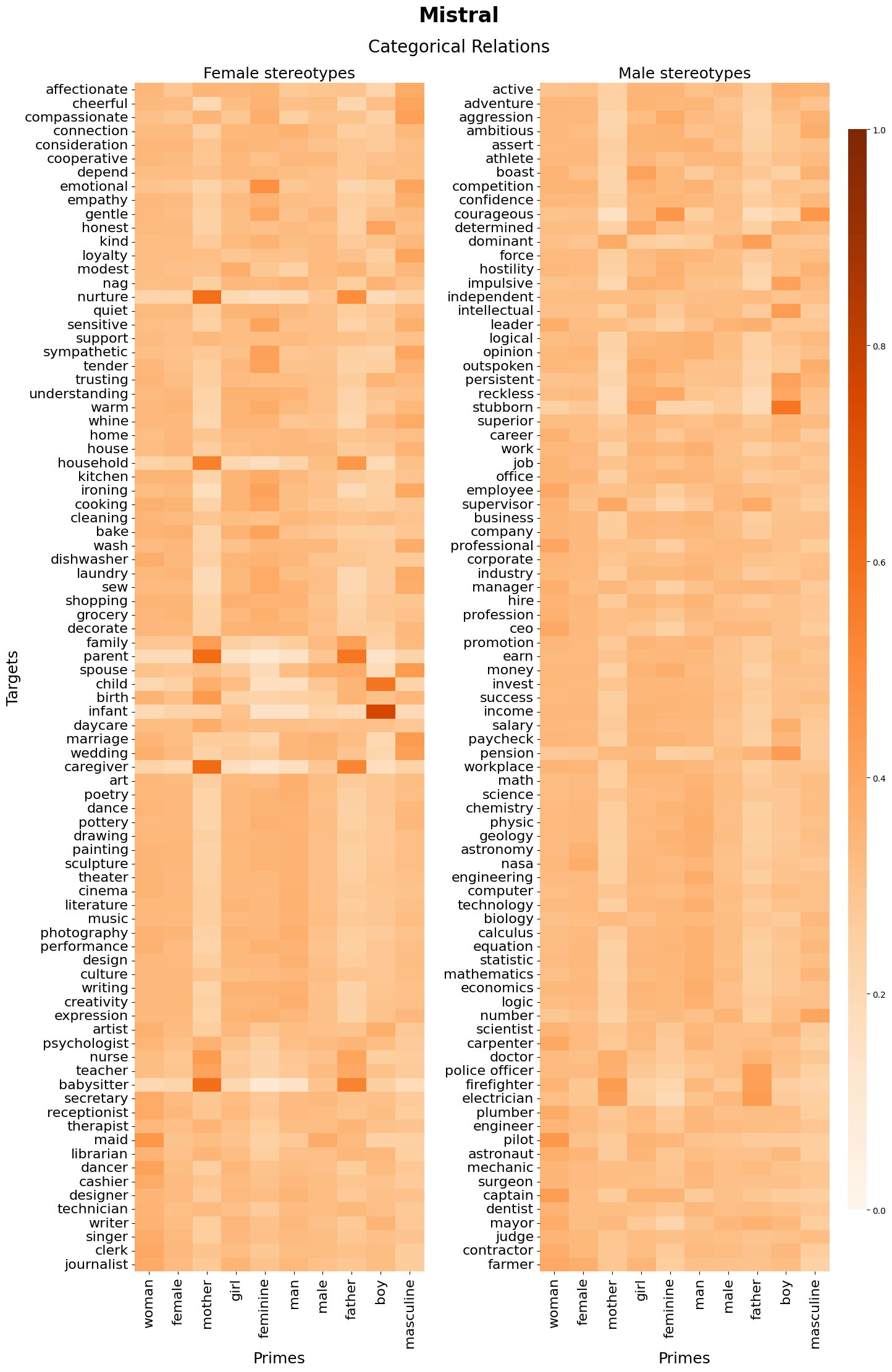}
\caption{\textbf{Normalized activation levels for the categorical relations layer in Mistral.} Normalized activation levels appear mostly homogeneous, but with several exceptions of prime-target pairs with high or low activation levels.}
\end{figure}

\begin{figure}[H]
\includegraphics[width=.8\linewidth]{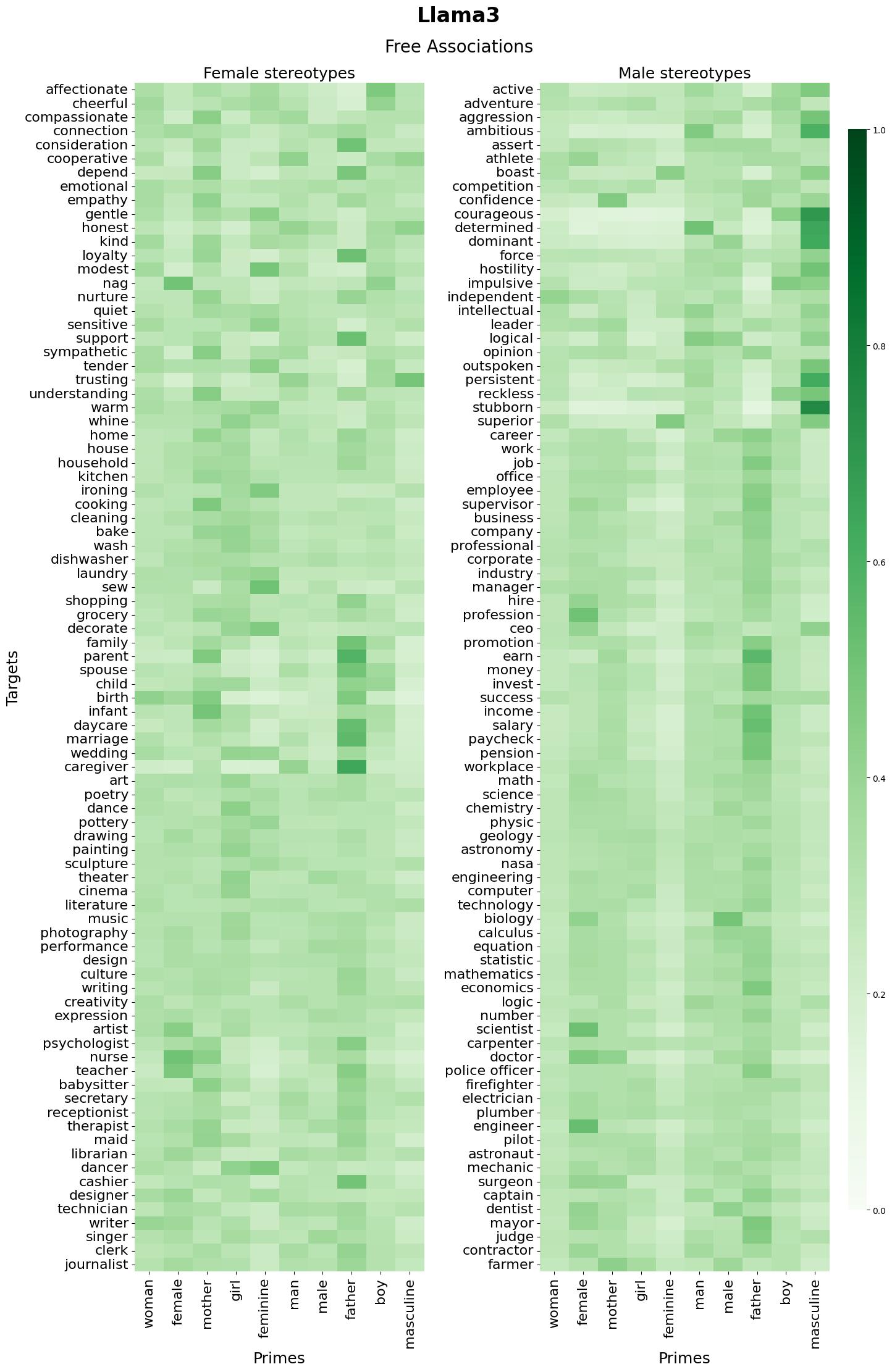}
\caption{\textbf{Normalized activation levels for the free associations layer in Llama3.} It appears that certain stereotype topics are particularly activated by certain primes (male-related traits are strongly activated by the prime \textit{masculine}.}
\end{figure}

\begin{figure}[H]
\includegraphics[width=.8\linewidth]{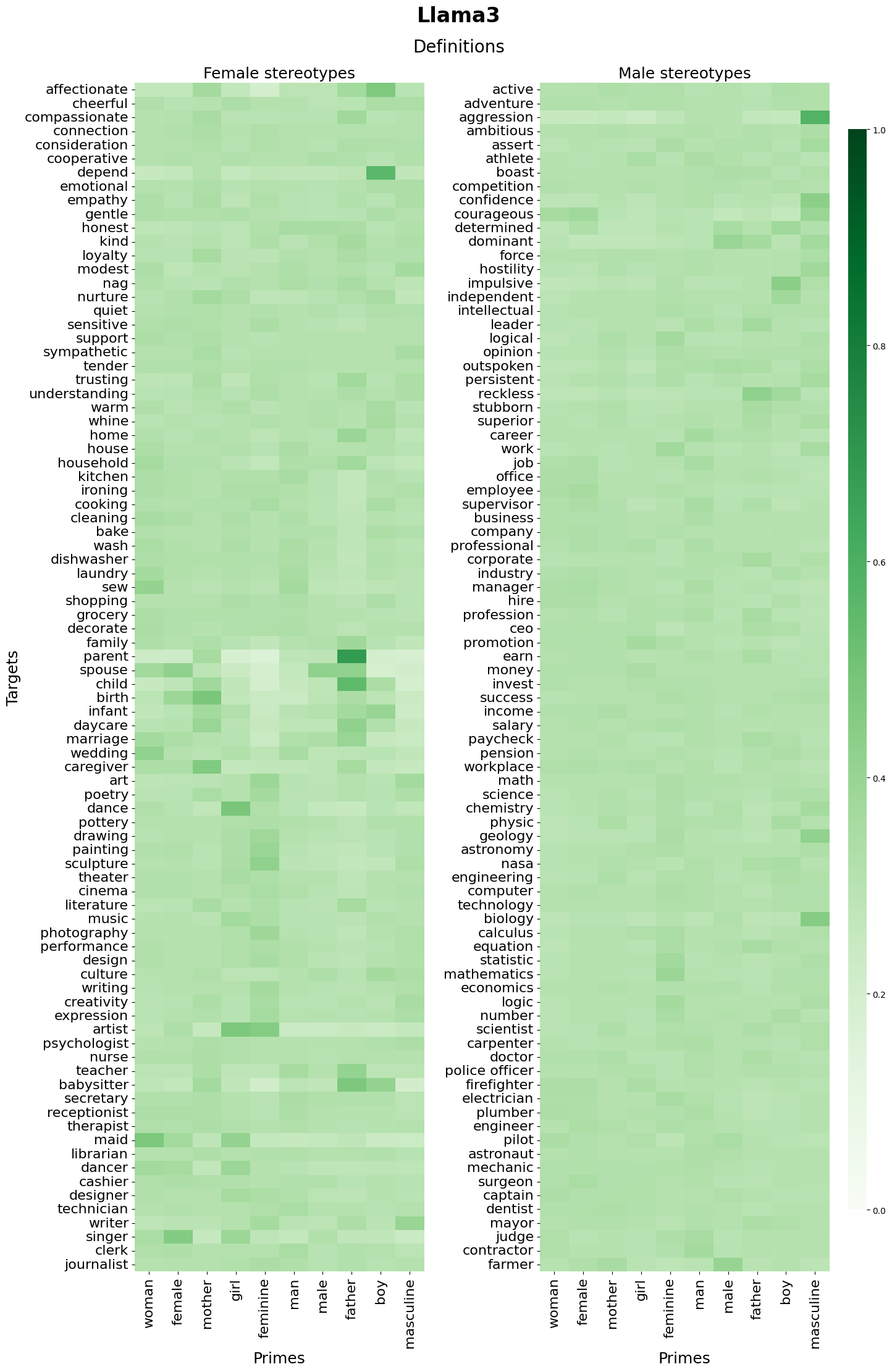}
\caption{\textbf{Normalized activation levels for the definitions layer in Llama3.} Normalized activation levels appear mostly homogeneous, with little variation across topics and layers, with the exception of a few prime-target pairs like \textit{father -- parent}.}
\end{figure}

\begin{figure}[H]
\includegraphics[width=.8\linewidth]{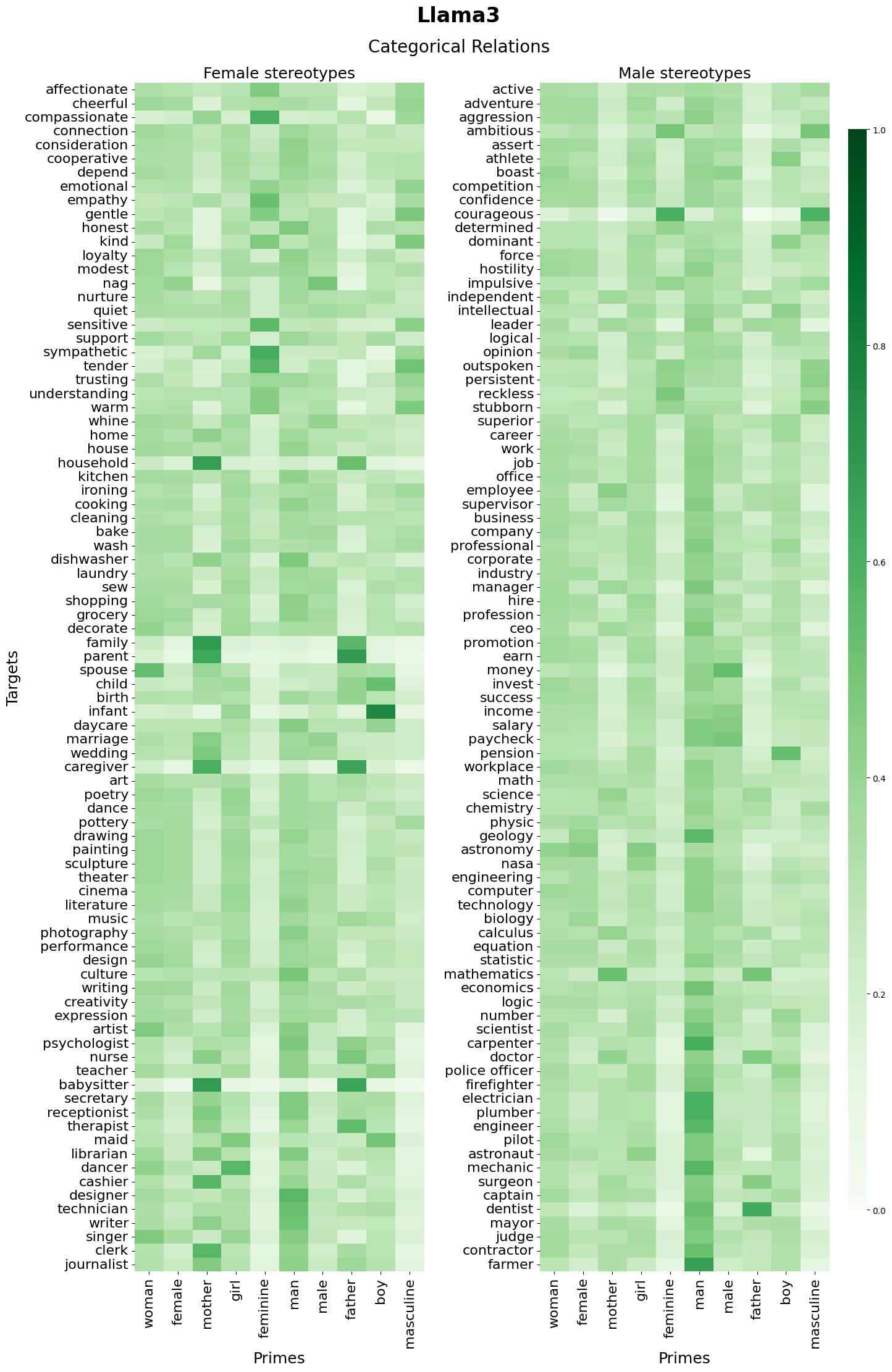}
\caption{\textbf{Normalized activation levels for the categorical relations layer in Llama3.} The heatmaps show large variations in the normalized activation levels of prime-target pairs, in contrast to the free associations and definitions layers.}
\end{figure}

\bibliographystyle{abbrv}
\bibliography{biblio}

\begin{thebibliography}{10}

\bibitem{abramski2025llm}
K.~Abramski, R.~Improta, G.~Rossetti, and M.~Stella.
\newblock The "llm world of words" english free association norms generated by large language models.
\newblock {\em Scientific data}, 12(1):803, 2025.

\bibitem{abramski2025word}
K.~Abramski, G.~Rossetti, and M.~Stella.
\newblock A word association network methodology for evaluating implicit biases in llms compared to humans.
\newblock {\em arXiv preprint arXiv:2510.24488}, 2025.

\bibitem{acerbi2023large}
A.~Acerbi and J.~M. Stubbersfield.
\newblock Large language models show human-like content biases in transmission chain experiments.
\newblock {\em Proceedings of the National Academy of Sciences}, 120(44):e2313790120, 2023.

\bibitem{agrawal2024can}
G.~Agrawal, T.~Kumarage, Z.~Alghamdi, and H.~Liu.
\newblock Can knowledge graphs reduce hallucinations in llms?: A survey.
\newblock In {\em Proceedings of the 2024 Conference of the North American Chapter of the Association for Computational Linguistics: Human Language Technologies (Volume 1: Long Papers)}, pages 3947--3960, 2024.

\bibitem{an2022learning}
H.~An, X.~Liu, and D.~Zhang.
\newblock Learning bias-reduced word embeddings using dictionary definitions.
\newblock In {\em Findings of the Association for Computational Linguistics: ACL 2022}, pages 1139--1152, 2022.

\bibitem{anderson1983spreading}
J.~R. Anderson.
\newblock A spreading activation theory of memory.
\newblock {\em Journal of verbal learning and verbal behavior}, 22(3):261--295, 1983.

\bibitem{anderson1998integrated}
J.~R. Anderson, D.~Bothell, C.~Lebiere, and M.~Matessa.
\newblock An integrated theory of list memory.
\newblock {\em Journal of Memory and Language}, 38(4):341--380, 1998.

\bibitem{anderson1997act}
J.~R. Anderson, M.~Matessa, and C.~Lebiere.
\newblock Act-r: A theory of higher level cognition and its relation to visual attention.
\newblock {\em Human--Computer Interaction}, 12(4):439--462, 1997.

\bibitem{bai2025explicitly}
X.~Bai, A.~Wang, I.~Sucholutsky, and T.~L. Griffiths.
\newblock Explicitly unbiased large language models still form biased associations.
\newblock {\em Proceedings of the National Academy of Sciences}, 122(8):e2416228122, 2025.

\bibitem{bargh1999unbearable}
J.~A. Bargh and T.~L. Chartrand.
\newblock The unbearable automaticity of being.
\newblock {\em American psychologist}, 54(7):462, 1999.

\bibitem{barr1981handbook}
A.~Barr, E.~A. Feigenbaum, and P.~R. Cohen.
\newblock {\em The handbook of artificial intelligence}, volume~1.
\newblock HeurisTech Press, 1981.

\bibitem{bender2021dangers}
E.~M. Bender, T.~Gebru, A.~McMillan-Major, and S.~Shmitchell.
\newblock On the dangers of stochastic parrots: Can language models be too big?
\newblock In {\em Proceedings of the 2021 ACM conference on fairness, accountability, and transparency}, pages 610--623, 2021.

\bibitem{boccaletti2014structure}
S.~Boccaletti, G.~Bianconi, R.~Criado, C.~I. Del~Genio, J.~G{\'o}mez-Gardenes, M.~Romance, I.~Sendina-Nadal, Z.~Wang, and M.~Zanin.
\newblock The structure and dynamics of multilayer networks.
\newblock {\em Physics reports}, 544(1):1--122, 2014.

\bibitem{brachman1979epistemological}
R.~J. Brachman.
\newblock On the epistemological status of semantic networks.
\newblock In {\em Associative networks}, pages 3--50. Elsevier, 1979.

\bibitem{brady2025dual}
O.~Brady, P.~Nulty, L.~Zhang, T.~E. Ward, and D.~P. McGovern.
\newblock Dual-process theory and decision-making in large language models.
\newblock {\em Nature Reviews Psychology}, pages 1--16, 2025.

\bibitem{bursell2021we}
M.~Bursell and F.~Olsson.
\newblock Do we need dual-process theory to understand implicit bias? a study of the nature of implicit bias against muslims.
\newblock {\em Poetics}, 87:101549, 2021.

\bibitem{castro2020contributions}
N.~Castro and C.~S. Siew.
\newblock Contributions of modern network science to the cognitive sciences: Revisiting research spirals of representation and process.
\newblock {\em Proceedings of the Royal Society A}, 476(2238):20190825, 2020.

\bibitem{citraro2023feature}
S.~Citraro, M.~S. Vitevitch, M.~Stella, and G.~Rossetti.
\newblock Feature-rich multiplex lexical networks reveal mental strategies of early language learning.
\newblock {\em Scientific Reports}, 13(1):1474, 2023.

\bibitem{collins1975spreading}
A.~M. Collins and E.~F. Loftus.
\newblock A spreading-activation theory of semantic processing.
\newblock {\em Psychological review}, 82(6):407, 1975.

\bibitem{collins1969retrieval}
A.~M. Collins and M.~R. Quillian.
\newblock Retrieval time from semantic memory.
\newblock {\em Journal of verbal learning and verbal behavior}, 8(2):240--247, 1969.

\bibitem{de2019small}
S.~De~Deyne, D.~J. Navarro, A.~Perfors, M.~Brysbaert, and G.~Storms.
\newblock The "small world of words" english word association norms for over 12,000 cue words.
\newblock {\em Behavior research methods}, 51(3):987--1006, 2019.

\bibitem{de2015structural}
M.~De~Domenico, V.~Nicosia, A.~Arenas, and V.~Latora.
\newblock Structural reducibility of multilayer networks.
\newblock {\em Nature communications}, 6(1):6864, 2015.

\bibitem{de2013mathematical}
M.~De~Domenico, A.~Sol{\'e}-Ribalta, E.~Cozzo, M.~Kivel{\"a}, Y.~Moreno, M.~A. Porter, S.~G{\'o}mez, and A.~Arenas.
\newblock Mathematical formulation of multilayer networks.
\newblock {\em Physical Review X}, 3(4):041022, 2013.

\bibitem{de2026cognitive}
E.~S. De~Duro, E.~Franchino, R.~Improta, G.~A. Veltri, and M.~Stella.
\newblock Cognitive networks identify ai biases on societal issues in large language models.
\newblock {\em EPJ Data Science}, 15(1):7, 2026.

\bibitem{de2021attitudes}
J.~De~Houwer, P.~Van~Dessel, and T.~Moran.
\newblock Attitudes as propositional representations.
\newblock {\em Trends in Cognitive Sciences}, 25(10):870--882, 2021.

\bibitem{de2024system}
J.~C. de~Winter, D.~Dodou, and Y.~B. Eisma.
\newblock System 2 thinking in openai's o1-preview model: Near-perfect performance on a mathematics exam.
\newblock {\em Computers}, 13(11):278, 2024.

\bibitem{devine1989stereotypes}
P.~G. Devine.
\newblock Stereotypes and prejudice: Their automatic and controlled components.
\newblock {\em Journal of personality and social psychology}, 56(1):5, 1989.

\bibitem{devine2012long}
P.~G. Devine, P.~S. Forscher, A.~J. Austin, and W.~T. Cox.
\newblock Long-term reduction in implicit race bias: A prejudice habit-breaking intervention.
\newblock {\em Journal of experimental social psychology}, 48(6):1267--1278, 2012.

\bibitem{evans2008dual}
J.~S.~B. Evans.
\newblock Dual-processing accounts of reasoning, judgment, and social cognition.
\newblock {\em Annu. Rev. Psychol.}, 59(1):255--278, 2008.

\bibitem{evans2010intuition}
J.~S.~B. Evans.
\newblock Intuition and reasoning: A dual-process perspective.
\newblock {\em Psychological Inquiry}, 21(4):313--326, 2010.

\bibitem{evans2011dual}
J.~S.~B. Evans.
\newblock Dual-process theories of reasoning: Contemporary issues and developmental applications.
\newblock {\em Developmental review}, 31(2-3):86--102, 2011.

\bibitem{evans2013dual}
J.~S.~B. Evans and K.~E. Stanovich.
\newblock Dual-process theories of higher cognition: Advancing the debate.
\newblock {\em Perspectives on psychological science}, 8(3):223--241, 2013.

\bibitem{fodor1975language}
J.~Fodor.
\newblock {\em The language of thought}.
\newblock Harvard University Press, 1975.

\bibitem{fodor1988connectionism}
J.~A. Fodor and Z.~W. Pylyshyn.
\newblock Connectionism and cognitive architecture: A critical analysis.
\newblock {\em Cognition}, 28(1-2):3--71, 1988.

\bibitem{garimella2021he}
A.~Garimella, A.~Amarnath, K.~Kumar, A.~P. Yalla, N.~Chhaya, B.~V. Srinivasan, et~al.
\newblock He is very intelligent, she is very beautiful? on mitigating social biases in language modelling and generation.
\newblock In {\em Findings of the association for computational linguistics: ACL-IJCNLP 2021}, pages 4534--4545, 2021.

\bibitem{gawronski2006associative}
B.~Gawronski and G.~V. Bodenhausen.
\newblock Associative and propositional processes in evaluation: an integrative review of implicit and explicit attitude change.
\newblock {\em Psychological bulletin}, 132(5):692, 2006.

\bibitem{gawronski2011associative}
B.~Gawronski and G.~V. Bodenhausen.
\newblock The associative--propositional evaluation model: Theory, evidence, and open questions.
\newblock {\em Advances in experimental social psychology}, 44:59--127, 2011.

\bibitem{greenwald1995implicit}
A.~G. Greenwald and M.~R. Banaji.
\newblock Implicit social cognition: attitudes, self-esteem, and stereotypes.
\newblock {\em Psychological review}, 102(1):4, 1995.

\bibitem{greenwald1998measuring}
A.~G. Greenwald, D.~E. McGhee, and J.~L. Schwartz.
\newblock Measuring individual differences in implicit cognition: the implicit association test.
\newblock {\em Journal of personality and social psychology}, 74(6):1464, 1998.

\bibitem{hagendorff2023human}
T.~Hagendorff, S.~Fabi, and M.~Kosinski.
\newblock Human-like intuitive behavior and reasoning biases emerged in large language models but disappeared in chatgpt.
\newblock {\em Nature Computational Science}, 3(10):833--838, 2023.

\bibitem{hills2022mind}
T.~T. Hills and Y.~N. Kenett.
\newblock Is the mind a network? maps, vehicles, and skyhooks in cognitive network science.
\newblock {\em Topics in Cognitive Science}, 14(1):189--208, 2022.

\bibitem{hutchison2013semantic}
K.~A. Hutchison, D.~A. Balota, J.~H. Neely, M.~J. Cortese, E.~R. Cohen-Shikora, C.-S. Tse, M.~J. Yap, J.~J. Bengson, D.~Niemeyer, and E.~Buchanan.
\newblock The semantic priming project.
\newblock {\em Behavior research methods}, 45:1099--1114, 2013.

\bibitem{kahneman2011thinking}
D.~Kahneman.
\newblock {\em Thinking, fast and slow}.
\newblock macmillan, 2011.

\bibitem{kamruzzaman2024prompting}
M.~Kamruzzaman and G.~L. Kim.
\newblock Prompting techniques for reducing social bias in llms through system 1 and system 2 cognitive processes.
\newblock {\em arXiv preprint arXiv:2404.17218}, 2024.

\bibitem{kaneko2021dictionary}
M.~Kaneko and D.~Bollegala.
\newblock Dictionary-based debiasing of pre-trained word embeddings.
\newblock {\em arXiv preprint arXiv:2101.09525}, 2021.

\bibitem{kintsch1998comprehension}
W.~Kintsch.
\newblock {\em Comprehension: A paradigm for cognition}.
\newblock Cambridge university press, 1998.

\bibitem{kivela2014multilayer}
M.~Kivel{\"a}, A.~Arenas, M.~Barthelemy, J.~P. Gleeson, Y.~Moreno, and M.~A. Porter.
\newblock Multilayer networks.
\newblock {\em Journal of complex networks}, 2(3):203--271, 2014.

\bibitem{kozima1996similarity}
H.~Kozima and T.~Furugori.
\newblock Similarity between words computed by spreading activation on an english dictionary.
\newblock {\em arXiv preprint cmp-lg/9601004}, 1996.

\bibitem{kumar2025detecting}
R.~Kumar, H.~Kumar, and K.~Shalini.
\newblock Detecting and mitigating bias in llms through knowledge graph-augmented training.
\newblock In {\em 2025 International Conference on Artificial Intelligence and Data Engineering (AIDE)}, pages 608--613. IEEE, 2025.

\bibitem{li2025system}
Z.-Z. Li, D.~Zhang, M.-L. Zhang, J.~Zhang, Z.~Liu, Y.~Yao, H.~Xu, J.~Zheng, P.-J. Wang, X.~Chen, et~al.
\newblock From system 1 to system 2: A survey of reasoning large language models.
\newblock {\em arXiv preprint arXiv:2502.17419}, 2025.

\bibitem{ma2024debiasing}
C.~Ma, T.~Zhao, and M.~Okumura.
\newblock Debiasing large language models with structured knowledge.
\newblock In {\em Findings of the Association for Computational Linguistics: ACL 2024}, pages 10274--10287, 2024.

\bibitem{mcrae2005semantic}
K.~McRae, G.~S. Cree, M.~S. Seidenberg, and C.~McNorgan.
\newblock Semantic feature production norms for a large set of living and nonliving things.
\newblock {\em Behavior research methods}, 37(4):547--559, 2005.

\bibitem{miller1995wordnet}
G.~A. Miller.
\newblock Wordnet: a lexical database for english.
\newblock {\em Communications of the ACM}, 38(11):39--41, 1995.

\bibitem{monteith1993self}
M.~J. Monteith.
\newblock Self-regulation of prejudiced responses: Implications for progress in prejudice-reduction efforts.
\newblock {\em Journal of personality and social psychology}, 65(3):469, 1993.

\bibitem{monteith2001taking}
M.~J. Monteith, C.~I. Voils, and L.~Ashburn-Nardo.
\newblock Taking a look underground: Detecting, interpreting, and reacting to implicit racial biases.
\newblock {\em Social Cognition}, 19(4):395--417, 2001.

\bibitem{moors2006problems}
A.~Moors and J.~De~Houwer.
\newblock Problems with dividing the realm of processes.
\newblock {\em Psychological Inquiry}, 17(3):199--204, 2006.

\bibitem{mruthyunjaya2023rethinking}
V.~Mruthyunjaya, P.~Pezeshkpour, E.~Hruschka, and N.~Bhutani.
\newblock Rethinking language models as symbolic knowledge graphs.
\newblock {\em arXiv preprint arXiv:2308.13676}, 2023.

\bibitem{nosek2007pervasiveness}
B.~A. Nosek, F.~L. Smyth, J.~J. Hansen, T.~Devos, N.~M. Lindner, K.~A. Ranganath, C.~T. Smith, K.~R. Olson, D.~Chugh, A.~G. Greenwald, et~al.
\newblock Pervasiveness and correlates of implicit attitudes and stereotypes.
\newblock {\em European review of social psychology}, 18(1):36--88, 2007.

\bibitem{pavlick2023symbols}
E.~Pavlick.
\newblock Symbols and grounding in large language models.
\newblock {\em Philosophical Transactions of the Royal Society A}, 381(2251):20220041, 2023.

\bibitem{payne2005conceptualizing}
B.~K. Payne.
\newblock Conceptualizing control in social cognition: how executive functioning modulates the expression of automatic stereotyping.
\newblock {\em Journal of personality and social psychology}, 89(4):488, 2005.

\bibitem{perugini2005predictive}
M.~Perugini.
\newblock Predictive models of implicit and explicit attitudes.
\newblock {\em British Journal of Social Psychology}, 44(1):29--45, 2005.

\bibitem{reder1980partial}
L.~M. Reder and J.~R. Anderson.
\newblock A partial resolution of the paradox of interference: The role of integrating knowledge.
\newblock {\em Cognitive Psychology}, 12(4):447--472, 1980.

\bibitem{rumelhart1986parallel}
D.~E. Rumelhart, J.~L. McClelland, P.~R. Group, et~al.
\newblock {\em Parallel distributed processing, volume 1: Explorations in the microstructure of cognition: Foundations}.
\newblock The MIT press, 1986.

\bibitem{siew2019spreadr}
C.~S. Siew.
\newblock spreadr: An r package to simulate spreading activation in a network.
\newblock {\em Behavior Research Methods}, 51(2):910--929, 2019.

\bibitem{siew2019cognitive}
C.~S. Siew, D.~U. Wulff, N.~M. Beckage, and Y.~N. Kenett.
\newblock Cognitive network science: A review of research on cognition through the lens of network representations, processes, and dynamics.
\newblock {\em Complexity}, 2019, 2019.

\bibitem{simon1971human}
H.~A. Simon and A.~Newell.
\newblock Human problem solving: The state of the theory in 1970.
\newblock {\em American psychologist}, 26(2):145, 1971.

\bibitem{sloman1996empirical}
S.~A. Sloman.
\newblock The empirical case for two systems of reasoning.
\newblock {\em Psychological bulletin}, 119(1):3, 1996.

\bibitem{smith1974structure}
E.~E. Smith, E.~J. Shoben, and L.~J. Rips.
\newblock Structure and process in semantic memory: A featural model for semantic decisions.
\newblock {\em Psychological review}, 81(3):214, 1974.

\bibitem{smith2000dual}
E.~R. Smith and J.~DeCoster.
\newblock Dual-process models in social and cognitive psychology: Conceptual integration and links to underlying memory systems.
\newblock {\em Personality and social psychology review}, 4(2):108--131, 2000.

\bibitem{smolensky1988proper}
P.~Smolensky.
\newblock On the proper treatment of connectionism.
\newblock {\em Behavioral and brain sciences}, 11(1):1--23, 1988.

\bibitem{sowa1983generating}
J.~F. Sowa.
\newblock Generating language from conceptual graphs.
\newblock In {\em Computational Linguistics}, pages 29--43. Elsevier, 1983.

\bibitem{stanovich1999rational}
K.~E. Stanovich.
\newblock {\em Who is rational?: Studies of individual differences in reasoning}.
\newblock Psychology Press, 1999.

\bibitem{stella2024cognitive}
M.~Stella, S.~Citraro, G.~Rossetti, D.~Marinazzo, Y.~N. Kenett, and M.~S. Vitevitch.
\newblock Cognitive modelling of concepts in the mental lexicon with multilayer networks: Insights, advancements, and future challenges.
\newblock {\em Psychonomic Bulletin \& Review}, 31(5):1981--2004, 2024.

\bibitem{steyvers2005large}
M.~Steyvers and J.~B. Tenenbaum.
\newblock The large-scale structure of semantic networks: Statistical analyses and a model of semantic growth.
\newblock {\em Cognitive science}, 29(1):41--78, 2005.

\bibitem{strack2004reflective}
F.~Strack and R.~Deutsch.
\newblock Reflective and impulsive determinants of social behavior.
\newblock {\em Personality and social psychology review}, 8(3):220--247, 2004.

\bibitem{vincent2016latent}
P.~Vincent-Lamarre, A.~B. Mass{\'e}, M.~Lopes, M.~Lord, O.~Marcotte, and S.~Harnad.
\newblock The latent structure of dictionaries.
\newblock {\em Topics in cognitive science}, 8(3):625--659, 2016.

\bibitem{wei2022emergent}
J.~Wei, Y.~Tay, R.~Bommasani, C.~Raffel, B.~Zoph, S.~Borgeaud, D.~Yogatama, M.~Bosma, D.~Zhou, D.~Metzler, et~al.
\newblock Emergent abilities of large language models.
\newblock {\em arXiv preprint arXiv:2206.07682}, 2022.

\bibitem{wei2022chain}
J.~Wei, X.~Wang, D.~Schuurmans, M.~Bosma, F.~Xia, E.~Chi, Q.~V. Le, D.~Zhou, et~al.
\newblock Chain-of-thought prompting elicits reasoning in large language models.
\newblock {\em Advances in Neural Information Processing Systems}, 35:24824--24837, 2022.

\bibitem{woods1975s}
W.~A. Woods.
\newblock What's in a link: Foundations for semantic networks.
\newblock In {\em Representation and understanding}, pages 35--82. Elsevier, 1975.

\bibitem{xiang2025towards}
V.~Xiang, C.~Snell, K.~Gandhi, A.~Albalak, A.~Singh, C.~Blagden, D.~Phung, R.~Rafailov, N.~Lile, D.~Mahan, et~al.
\newblock Towards system 2 reasoning in llms: Learning how to think with meta chain-of-thought.
\newblock {\em arXiv preprint arXiv:2501.04682}, 2025.

\bibitem{yang2025llm2}
C.~Yang, C.~Shi, S.~Li, B.~Shui, Y.~Yang, and W.~Lam.
\newblock Llm2: Let large language models harness system 2 reasoning.
\newblock In {\em Proceedings of the 2025 Conference of the Nations of the Americas Chapter of the Association for Computational Linguistics: Human Language Technologies (Volume 2: Short Papers)}, pages 168--177, 2025.

\bibitem{yax2024studying}
N.~Yax, H.~Anll{\'o}, and S.~Palminteri.
\newblock Studying and improving reasoning in humans and machines.
\newblock {\em Communications Psychology}, 2(1):51, 2024.

\bibitem{zhao2024comparative}
Y.~Zhao, B.~Wang, Y.~Wang, D.~Zhao, X.~Jin, J.~Zhang, R.~He, and Y.~Hou.
\newblock A comparative study of explicit and implicit gender biases in large language models via self-evaluation.
\newblock In {\em Proceedings of the 2024 Joint International Conference on Computational Linguistics, Language Resources and Evaluation (LREC-COLING 2024)}, pages 186--198, 2024.

\bibitem{zheng2023does}
S.~Zheng, J.~Huang, and K.~C.-C. Chang.
\newblock Why does chatgpt fall short in answering questions faithfully.
\newblock {\em arXiv preprint arXiv:2304.10513}, 2, 2023.

\bibitem{ziabari2025reasoning}
A.~S. Ziabari, N.~Ghazizadeh, Z.~Sourati, F.~Karimi-Malekabadi, P.~Piray, and M.~Dehghani.
\newblock Reasoning on a spectrum: Aligning llms to system 1 and system 2 thinking.
\newblock {\em arXiv preprint arXiv:2502.12470}, 2025.

\end{thebibliography}

\end{document}